\documentclass[preprint,3p]{elsarticle}

\usepackage{amssymb}

\usepackage{graphicx}
\usepackage{caption}
\usepackage{subcaption}
\usepackage{mdframed}
\usepackage{xspace}
\usepackage{mathtools}
\usepackage{booktabs}
\usepackage{url}
\usepackage[T1]{fontenc}

\journal{Elsevier for possible publication.}

\newcommand{\ie}{\emph{i.e.,}\xspace}
\newcommand{\eg}{\emph{e.g.,}\xspace}

\begin{document}

\begin{frontmatter}



\title{A Hybrid Ensemble Feature Selection Design for Candidate Biomarkers Discovery from Transcriptome Profiles}


\author[1,2]{Felipe Colombelli}
\ead{fcolombelli@inf.ufrgs.br}

\author[2,3,4,5]{Thayne Woycinck Kowalski}
\ead{tkowalski@hcpa.edu.br}
    
\author[1,2]{Mariana Recamonde-Mendoza\corref{cor1}}
\ead{mrmendoza@inf.ufrgs.br}

\cortext[cor1]{Corresponding author}
\address[1]{Institute of Informatics, Universidade Federal do Rio Grande do Sul, Porto Alegre, RS, Brazil}
\address[2]{Bioinformatics Core, Hospital de Clínicas de Porto Alegre, Porto Alegre, RS, Brazil}
\address[3]{Post-Graduation Program in Genetics and Molecular Biology, Genetics Department, Universidade Federal do Rio Grande do Sul, Porto Alegre, RS, Brazil}
\address[4]{Laboratory of Genomic Medicine, Center of Experimental Research, Hospital de Clínicas de Porto Alegre, Porto Alegre, RS, Brazil}
\address[5]{Centro Universitário CESUCA, Cachoeirinha, RS, Brazil}

\begin{abstract} 

The discovery of disease biomarkers from gene expression data has been greatly advanced by feature selection (FS) methods, especially using ensemble FS (EFS) strategies with perturbation at the data level (\ie homogeneous, Hom-EFS) or method level (\ie heterogeneous, Het-EFS). Here we proposed a Hybrid EFS (Hyb-EFS) design that explores both types of perturbation to improve the stability and the predictive power of candidate biomarkers. With this, Hyb-EFS aims to disrupt associations of good performance with a single dataset, single algorithm, or a specific combination of both, which is particularly interesting for better reproducibility of genomic biomarkers. We investigated the adequacy of our approach for microarray data related to four types of cancer, carrying out an extensive comparison with other ensemble and single FS approaches. Five FS methods were used in our experiments: Wx, Symmetrical Uncertainty (SU), Gain Ratio (GR), Characteristic Direction (GeoDE), and ReliefF. We observed that the Hyb-EFS and Het-EFS approaches attenuated the large performance variation observed for most single FS and Hom-EFS across distinct datasets. Also, the Hyb-EFS improved upon the stability of the Het-EFS within our domain. Comparing the Hyb-EFS and Het-EFS composed of the top-performing selectors (Wx, GR, and SU), our hybrid approach surpassed the equivalent heterogeneous design and the best Hom-EFS (Hom-Wx). Interestingly, the rankings produced by our Hyb-EFS reached greater biological plausibility, with a notably high enrichment for cancer-related genes and pathways. Thus, our experiments suggest the potential of the proposed Hybrid EFS design in discovering candidate biomarkers from microarray data. Finally, we provide an open-source framework to support similar analyses in other domains, both as a user-friendly application and a plain Python package.
\end{abstract}

\begin{keyword}
feature selection \sep ensemble learning \sep biomarkers discovery \sep microarray \sep bioinformatics \sep high-dimensional data
\end{keyword}
\end{frontmatter}

\section{Introduction}
\label{intro}

The rise of bioinformatics in conjunction with new technologies for high-throughput analysis of biological systems triggered an exponential increase of data collection to study complex diseases at the molecular level \citep{gauthier2019brief}, leading to a vast amount of publicly available omics data. The characteristics and scientific impact of omics data have been extensively studied \citep{berger2013computational,quinn2019field,perez2019quantifying} and their integration is also a subject of growing interest \citep{huang2017more,manzoni2018genome}. 

One of the possibilities offered by the analysis of omics data such as transcriptome is discovering genes that may serve as biomarkers \citep{vittrant2020identification}. 
This finding is not only essential for understanding the mechanisms by which the disease operates but also allows the molecular characterization of the disease that may be further applied to stratify patients into well-defined groups according to the correlation with their clinical data. Therefore, biomarkers are of special interest to precision medicine provided their great utility for early diagnosis, anticipating adverse outcomes, and as new therapeutic targets \citep{Liu2019,karley2011biomarker}.

Machine learning (ML) algorithms have been successfully studied as analytical approaches for biomarkers discovery \citep{ledesma2021advancements,Zhang+2021}. Selecting candidate biomarkers from large-scale gene expression data can be directly translated to a classical and extensively investigated ML task called Feature Selection (FS) \citep{hira2015review}. The goal of FS is to reduce data dimensionality by finding the smallest subset of features that can correctly discriminate classes. In this particular domain, genes are interpreted as features and classes may be related to samples' diagnosis (\eg tumor or non-tumor), disease staging, response to therapy, among others. Although FS may be applied in a supervised, unsupervised, and semi-supervised fashion \citep{Ang+2015}, here, our focus is on supervised FS for binary classification tasks.

In the context of biomarkers discovery, selected candidate genes should have good discriminating power and be stable across data variations. Stability refers to observing consistent findings, \ie{ }identifying the same relevant features, when the FS process is repeated multiple times with data perturbation.  According to \citet{He&Yu2010}, stability is a good indicator of biomarker reproducibility, which is a required property for clinically useful biomarkers. Nonetheless, FS methods are often susceptible to variations in the training set, and distinct FS methods may extract different hypotheses from the same data \cite{Pes2020}. These issues make it challenging to translate the findings into clinical practice, as well as selecting the best FS a non-trivial task since the performance is highly data-dependent.

In order to mitigate these issues and achieve greater robustness in the FS process, Ensemble Feature Selection (EFS) has been actively investigated in several domains \citep{Bolon-Canedo2019,ali2018uefs}, including biomarkers discovery \cite{Zhang+2021}. Ensemble learning aims at combining multiple models to solve a ML task, based on the assumption that the diversity among hypotheses may improve overall performance compared to any single model.  When models are derived from the same algorithm, the ensemble is homogeneous; otherwise, it is called heterogeneous.

Considering the applications of EFS to biomarkers discovery, although diversity is commonly injected either by manipulating data input (homogeneous approaches) \citep{saeys2008robust,abeel2010robust} or applying distinct algorithms (heterogeneous approaches)  \citep{lopez2019automatic}, both strategies could be combined to boost diversity.  Hybrid EFS presents the advantage of introducing a much greater level of disruption into the FS process, which not only gives more confidence in the findings when stability between results is observed, but also a greater chance that stability is not tied to a single data composition or single algorithm, nor to a particular combination of both. This is of special interest in biomarkers discovery, although still underexplored in this domain \citep{zhang2019ensemble,dittman2012comparing}.

This work proposes a hybrid EFS approach for biomarkers discovery in which the bootstrap sampling technique is combined with multiple FS algorithms. The aggregation is performed in two steps. First, function perturbation is applied in each bootstrap and their rankings are aggregated, generating a sample-based consensus. Second, the rankings obtained for all bootstraps are aggregated to generate the final ensemble output, which represents the best consensus among distinct data and function perturbations. 
We perform an extensive evaluation of our approach against several base FS methods, including traditional ones and recently proposed methods \citep{park2019wx}, as well as homogeneous and heterogeneous EFS. Both predictive potential and stability are assessed and compared among the distinct approaches for four types of cancer (\ie breast, liver, lung, and pancreas), resulting in a broad investigation of EFS's potential and eventual limitations for dissecting transcriptome data in the search for new candidate biomarkers.

We note, however, that our approach is domain-independent and may be easily applied to other binary classification tasks. To facilitate its use in other problems, we implemented the proposed hybrid EFS approach, as well the homogeneous and heterogeneous EFS variations, as two open-source tools: a Python package, \texttt{efs-assembler}, containing functions to set up and run EFS experiments, and a graphical user interface solution, \texttt{BioSelector}, that allows easy configuration and evaluation of EFS methods for user informed datasets. More details about these tools are given in the text.

Altogether, our results aim at advancing the exploration of EFS methods, especially in its hybrid conception, to improve the stability and thus the clinical utility of candidate biomarkers extracted from transcriptome data. Whereas the majority of previous related studies that focused on gene expression data analysis miss the discussion of the biological plausibility of their findings, here, we carry out this analysis in addition to stability and performance assessment. We compared our findings with known cancer-related genes from robust databases, and we also used typical bioinformatics analysis to interrogate the functional role of genes top-ranked by our approach. We observed that many cancer-related genes and pathways are properly prioritized by our approach, which indicates that the proposed hybrid EFS has a great potential to suggest coherent candidates for cancer gene biomarkers to guide further experimental studies.

This paper is organized as follows. In Section~\ref{rel-works}, we provide an overview of related works on EFS, with emphasis on the domain of gene expression data analysis, and in Section~\ref{background} we present the theoretical background of our work. Sections \ref{method} to  \ref{results} describe our method, our experiments, and the results obtained. Section~\ref{sec:biologicalPlausibility} discusses our findings from a biological perspective, while Section~\ref{tools} provides details about the implemented tools made freely available for the scientific community. Finally, Section~\ref{conclusion} presents our conclusions.

\section{Related Works}
\label{rel-works}

By interpreting the problem of biomarker discovery from large-scale gene expression data as a feature selection task in ML, significant achievements have been reported in the scientific literature. Here, we concentrate on revising the main efforts related to EFS strategies within this domain.
\citet{saeys2008robust} and \citet{abeel2010robust}
showed early promising results with a homogeneous EFS for biomarker discovery using as a basis the SVM-Recursive Feature Elimination (RFE) algorithm, increasing both robustness and classification performance. \citet{lopez2019automatic} demonstrated the power of their heterogeneous EFS strategy for identifying miRNA biomarkers from cancers expression data. Their method outperformed widely used FS techniques like RFE, Univariate Feature Selection, and Genetic Algorithms when analyzing the accuracy outcomes. 

\citet{seijo2017ensemble} carried out an extensive analysis with thousands of experiments in different domain datasets, with different aggregation methods and different threshold approaches, demonstrating that the heterogeneous EFS was the most reliable FS strategy across multiple domain data regarding test error rate. Their analysis strongly contributed to the sense that, even though a particular FS algorithm could perform better for specific datasets and experiment configurations, in most cases (89.47\%), the heterogeneous EFS technique was not significantly worse than the best method. None of the single FS methods tested could achieve comparable reliability for the multiple experimented contexts. The same study pointed to the microarray data as the most susceptible to performance variations and the homogeneous EFS as a better strategy to tackle the FS process on it. \citet{Pes2020} investigated some configurations of a homogeneous EFS against its single FS algorithms and found that less stable algorithms benefit more from their ensemble version.

Other works, not necessarily investigating the domain of gene expression data, have also reinforced the evidence pointing to improvements of the EFS technique over the simple FS method. In \citet{ali2018uefs}, for example, a heterogeneous EFS composed only of filter methods was proposed. Their method consistently improved the non-ensemble versions in terms of predictive performance for various datasets, both textual and non-textual ones. \citet{das2020empirical} evaluated a heterogeneous EFS for Distributed Denial of Service (DDoS) attack detection and found an overall improvement in performance for the ensemble over the base FS algorithms.

Regarding the use of hybrid EFS approaches, in \citep{zhang2019ensemble} and \citep{chiew2019new}, authors proposed hybrid strategies that firstly aggregate rankings generated by data perturbation, and secondly combine results obtained by function perturbation – the opposed aggregation order explored by our design. 
\citet{zhang2019ensemble} described a general framework that was evaluated using gene expression data, but their primary goal was not to discover disease biomarkers. Additionally, the authors did not compare their hybrid EFS method with homogeneous versions of the base algorithms. Their findings suggest the high potential of the hybrid approach, greatly improving the heterogeneous setup in terms of stability and being one of the most stable methods in terms of predictive performance across different datasets. 

\citet{chiew2019new} focused their analysis on the spam detection problem, which has a much lower dimensionality (48 features) than our domain of interest. Additionally, the authors did not compare heterogeneous and homogeneous EFS approaches or assessed the proposed solution's stability. The focus of the evaluations was solely on the reduction of dimensionality without compromising the accuracy of the classifier. They showed that their hybrid ensemble method could deliver similar accuracy to that provided by the full dimensionality and was the method that selected the least number of features using their automatic threshold selection function. 

\citet{dittman2012comparing}, on the other hand, proposed a hybrid EFS design that applies one FS algorithm for its own sampled bootstrap bag, followed by a single aggregation process. Their findings, evaluated on microarray datasets, suggest greater predictive performance for the ensemble approaches using function perturbation when compared to the homogeneous EFS setup. However, not much insightful evidence was found in the performance comparisons made for the heterogeneous against the hybrid ensemble method.

\section{Background} 
\label{background}

\subsection{Feature selection}
\label{fs-methods}

Feature selection is commonly applied as a dimensionality reduction technique in ML, where the goal is to choose a subset of relevant features from the original set of features according to some pre-defined criteria. Several strategies have been proposed in the literature, as reviewed by \citet{Ang+2015}. Although other categorizations may apply, FS methods may be broadly divided into ranker and non-ranker approaches \cite{Bolon-Canedo2019}. Ranker methods employ some criterion to score each feature and provide a ranking of features relevance, based on which a subset of features may be selected by applying a threshold. In contrast, non-rankers return the subset of selected features without ranking or scoring information for unselected features \cite{Rokach+2007}. In this work, we are particularly interested in ranker methods since our final goal is to aggregate multiple feature relevance rankings using an ensemble strategy to reach more robust results.

Five FS methods were adopted in this work. The selection prioritized methods that demonstrated good performance in previous works, and that are both computationally efficient and independent of the classification algorithm. In addition, we aimed at including univariate and multivariate techniques, contemplating classic and more modern FS methods. We note that special consideration was also given to methods previously applied to omics data.

\begin{itemize}

  \item \textbf{Gain Ratio (GR)}
    applies the entropy concept to derive a ranking with the most explanatory features with respect to the class. It improves the usage of Information Gain (IG) alone by employing a normalizing factor that penalizes features resulting in many subcategories, which could be delivering high information gain and, thus, deceiving about the informative potential \citep{quinlan1986induction}.
  
  \item \textbf{Symmetrical Uncertainty (SU)} also relies in the notion of entropy values to build an feature relevance ranking. Essentially, it differs from GR by applying a different way to normalize the IG \citep{bommert2020benchmark}.

  \item \textbf{ReliefF} is an improvement of Relief algorithm, certainly one of the most popular FS methods. Relief scores each feature based on their differences between near instance pairs. Features scores are decreased if their values differ between nearby instances of the same class (a 'miss') and increase otherwise (a 'hit'). ReliefF strengthens the original Relief algorithm by averaging k hits and misses, hence delivering reliable probability approximation \citep{kononenko1994estimating,kononenko1997overcoming}.

  \item \textbf{Characteristic Direction (GeoDE)} is a recent FS technique proposed by \citet{clark2014characteristic} for the domain of gene expression data. The method takes concepts from linear classification, using the orientation of the separating hyperplane to infer the relative contribution of each component (\ie gene) to the total differential expression among classes. Thus, each gene receives a score reflecting its relative significance to discriminate among groups, which may be used to rank genes. 
  
  \item \textbf{Wx} is a neural network-based FS algorithm designed for transcriptome data in a very recent study \cite{park2019wx}. The algorithm trains an one-layer neural network based on softmax regression 
  and uses the model parameters learned from the training data for building an importance ranking using the features' discriminative index score.
\end{itemize}

\subsection{Ensemble feature selection}
\label{efs-method}
The idea behind EFS methods is to generate and aggregate diverse opinions about the importance of each feature. In other words, for the scope of this study, to produce different importance rankings. The ensemble paradigm advocates that a hidden knowledge about the importance of the features may emerge from the aggregation of different opinions in one final decision. This hidden knowledge can be seen as expanding the algorithms' search space, including new, possibly better, local optimum points. The aggregated opinion often overcomes single expert opinions in a phenomenon called Wisdom of the Crowds \citep{surowiecki2005wisdom,marbach2012wisdom}. As classical ensemble learning approaches, EFS involves two main steps: creating a set of diverse selectors (ensemble's base methods) and aggregating their results into a single final decision.

\subsubsection{Generation of ensemble base selectors}
\begin{figure*}[!ht]
\centering
  \begin{subfigure}[b]{0.48\textwidth}
    \includegraphics[width=\textwidth]{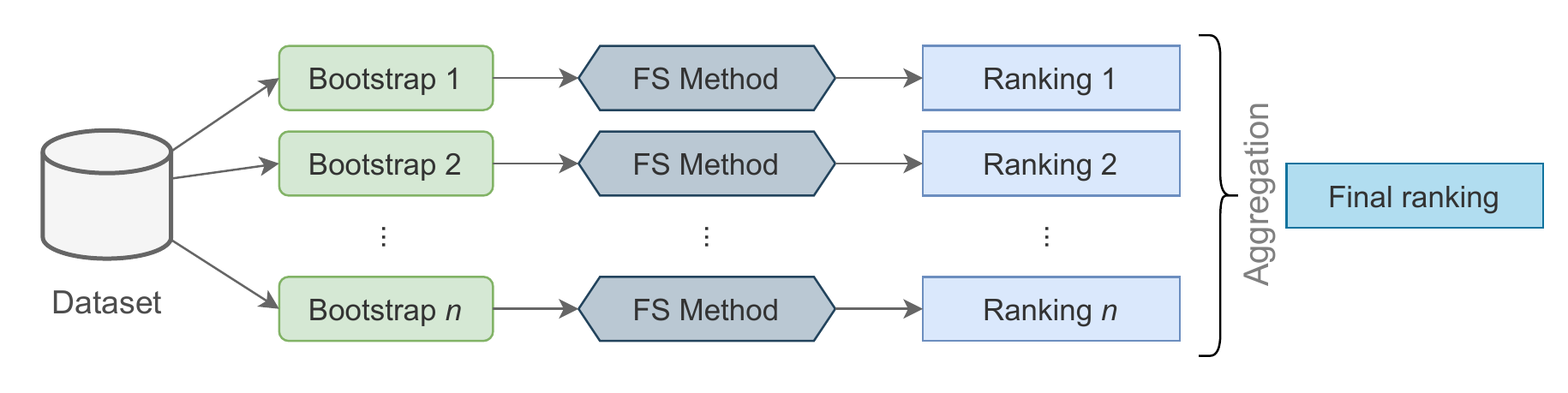}
    \caption{Homogeneous Ensemble.}
    \label{fig:ens-hom}
  \end{subfigure}
  \quad
  \begin{subfigure}[b]{0.48\textwidth}
    \includegraphics[width=\textwidth]{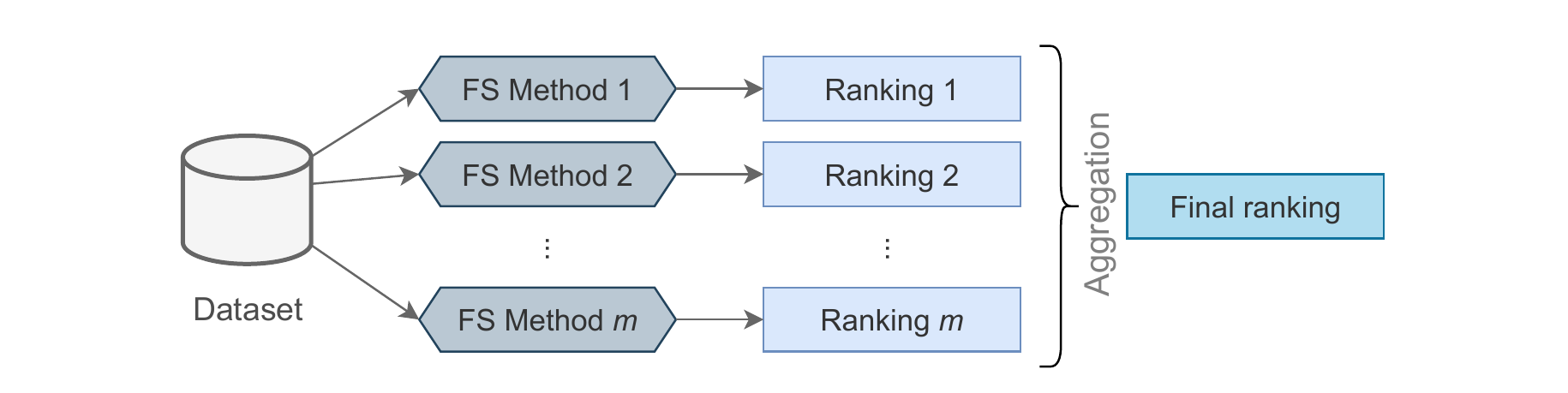}
    \caption{Heterogeneous Ensemble.}
    \label{fig:ens-het}
  \end{subfigure}
  \caption{Common used ensemble feature selection designs: a) homogeneous approaches explore diversity at data level, whereas b) heterogeneous approaches explore diversity among algorithms used as base selectors.}
  \label{fig:ens-types}
\end{figure*}

A critical part of any ensemble-based method is the diversity among opinions. In FS, the ensemble diversity is usually introduced using one of the following approaches: data perturbation or function perturbation, as compared in Figure \ref{fig:ens-types}.

Data perturbation (Figure \ref{fig:ens-hom}) applies the same FS method in \textit{n} random bootstrap samples drawn from the original data. This EFS design is referred to as Homogeneous EFS (Hom-EFS). The data perturbation approach has  improved the FS process's stability predominantly \citep{saeys2008robust,abeel2010robust,Pes2020}.

Function perturbation  (Figure \ref{fig:ens-het}) applies different FS methods to the original data, generating a Heterogeneous EFS (Het-EFS). The function perturbation approach has shown improvements towards the predictive potential of the FS process \citep{lopez2019automatic,ali2018uefs,seijo2017ensemble} and is currently the most common type of EFS strategy in literature \cite{Bolon-Canedo2019}.

\subsubsection{Rankings aggregation}
The second step in ensemble learning, including EFS, is to aggregate the diverse opinions into a single final decision. For the specific task of FS, we may consider that the final goal is to find a consensus ranking of features relevance by merging the knowledge brought by each individual ranking in the ensemble. Combining the multiple rankings generated in the first step into a single list of features ranked by importance is usually modeled as a rank aggregation problem, for which several methods have been developed \cite{Li2019}.

Simple arithmetic combination methods have been recurrently applied in related works \cite{Bolon-Canedo2019}. These include using operations such as mean, median, or minimum to summarize the position of each feature based on the analysis of its position in the multiple rankings, which are ordered in a decreasing order of importance. In the mean aggregation, for instance, the consensus position is the average position among all rankings, whereas in the minimum aggregation, each feature has its final position assigned according to the minimum (best) position that it has achieved among all rankings. Previous works also proposed the use of Borda Count, a very popular voting approach, as the aggregation method \cite{Drotar+2019}. Each base selector assigns $p$ points to a given feature, which corresponds to the number of features ranked below it in the selector's ranking. The aggregated score for each feature is computed as the average number of points assigned by all selectors and is used to generate the final consensus ranking. Both the original Borda count and the weighted Borda variant have shown good performance for EFS in previous works \cite{Drotar+2019}.

It is important to note that more sophisticated aggregation methods have also been proposed, aiming at improving the final ranking and avoiding any important loss of information. For instance, Robust Rank Aggregation proposes to compare the actual ranking to a null model based on random ordering of the rankings, from which p-values are extracted and used for reordering the features \cite{Kolde+2012}. Nonetheless, approaches that present higher computational complexity should be considered with cautious since the ensemble strategy per se presents high computational costs.

\subsection{Evaluation metrics}
\label{metrics}
There is a growing concern regarding dimensionality reduction in ML problems. Not only this kind of process reduces computational time for training classification models, but also tends to improve the model's predictive performance. Therefore, while reducing the features of a given dataset, it is interesting to analyze the predictive performance a particular classifier can achieve using only the selected features. Here, we refer to this criterion as the \textit{predictive potential} of a FS method. 

When there is a special interest in knowledge discovery from data, an additional evaluation criterion is the extent to which the FS method can choose mostly the same features while confronting different portions of the data \cite{Pes2020}. This indicates method's robustness with respect to changes in the input data. For the domain covered in the present paper, stability is a crucial characteristic as it is intrinsically associated to biomarkers reproducibility - a key property for establishing their clinical utility \cite{He&Yu2010}. Nonetheless, many related works in the field have neglected the importance of stability for designing EFS methods in the discovery of candidate biomarkers. In what follows we review common evaluation metrics used to assess predictive performance and stability of FS methods.

\subsubsection{Predictive potential}

Many different metrics are available to assess performance of classification models, all of which may be adopted to evaluate the predictive potential of FS results. In this scenario, a pre-defined learning algorithm (\eg{ } Support Vector Machine, Random Forest, Gradient Boosting Classifier, among others) is used to train a classifier with the distinct features subsets generated during the FS process, and considered as the predictive potential of the underlying selected features. Since the accuracy, \ie percentage of correct predictions, may be deceptive in scenarios with imbalanced classes, which is usually the case in disease-related omics data, other metrics are needed. 

The area under the ROC curve (ROC AUC) score is a commonly applied metric. The ROC curve shows the performance of a classification model at distinct thresholds settings, by plotting the corresponding True Positive Rate (TPR, in the y-axis) and the False Positive Rate (FPR, in the x-axis). This score may be interpreted as the probability that the model ranks a random positive instance more highly than a random negative instance. Therefore, a higher ROC AUC score indicates a better classification model.

\citet{Saito&Rehmsmeier2015} suggested that ROC AUC is not sensitive to different ratios of positives and negatives examples, owing to a wrong interpretation of models' specificity. Thus, the use of ROC AUC requires special caution when used with imbalanced datasets, especially when the positive class represents the minority one. In this scenario, the Precision-Recall (PR) curve, which shows precision values for corresponding recall (\ie sensitivity) values, better express the susceptibility of classifiers to class imbalance. Similar to the ROC AUC score, the area under the PR curve (PR AUC) summarizes the model's performance, where values closer to one indicate better classifiers. As \citet{Cook&Ramadas2020} states, the ROC curve is preferred when we care more about identifying a high percentage of the positives, whereas the PR AUC ensures that the positive predicted instances are mainly positive. Here we employ both metrics as they provide somewhat complementary information about model's predictive potential.

\subsubsection{Stability}

Let $A$ and $B$ be subsets of features selected from the same original set with cardinality $n$. Assume also that $|A|=|B|=k$ and that $r=|A \cap B|$. The consistency index $I_C$ between A and B \cite{kuncheva2007stability} is given by Equation \ref{equation-ic}.

\begin{equation}\label{equation-ic}
    I_C = \frac{rn - k^2}{k(n-k)}
\end{equation}

The Stability Index proposed by \citet{kuncheva2007stability}, here referred to as the Kuncheva Index (KI), is defined when there are multiple sets of selected features $\mathcal{A} = \{S_1,S_2,...,S_N\}$, for a given set size, $k$. The KI is simply the average of all pairwise consistency indexes, as shown in Equation \ref{equation-ki}.

\begin{equation}\label{equation-ki}
    KI(\mathcal{A}(N)) = \frac{2}{N(N-1)} \sum_{i=1}^{N-1} \sum_{j=i+1}^{N} I_C(S_i(k), S_j(k))
\end{equation}

The KI stability metric \citep{kuncheva2007stability} has been shown to bring theoretical and empirical improvements when compared to similarity index \citep{kalousis2005stability} and relative Hamming distance \citep{dunne2002solutions} in FS experiments. \citet{kuncheva2007stability} proposes that a high stability is achieved when the calculated KI is above 0.5.

\section{Proposed approach: a Hybrid Ensemble Feature Selection Design}
\label{method}

\begin{figure*}[!ht]
\centering
    \includegraphics[width=0.77\textwidth]{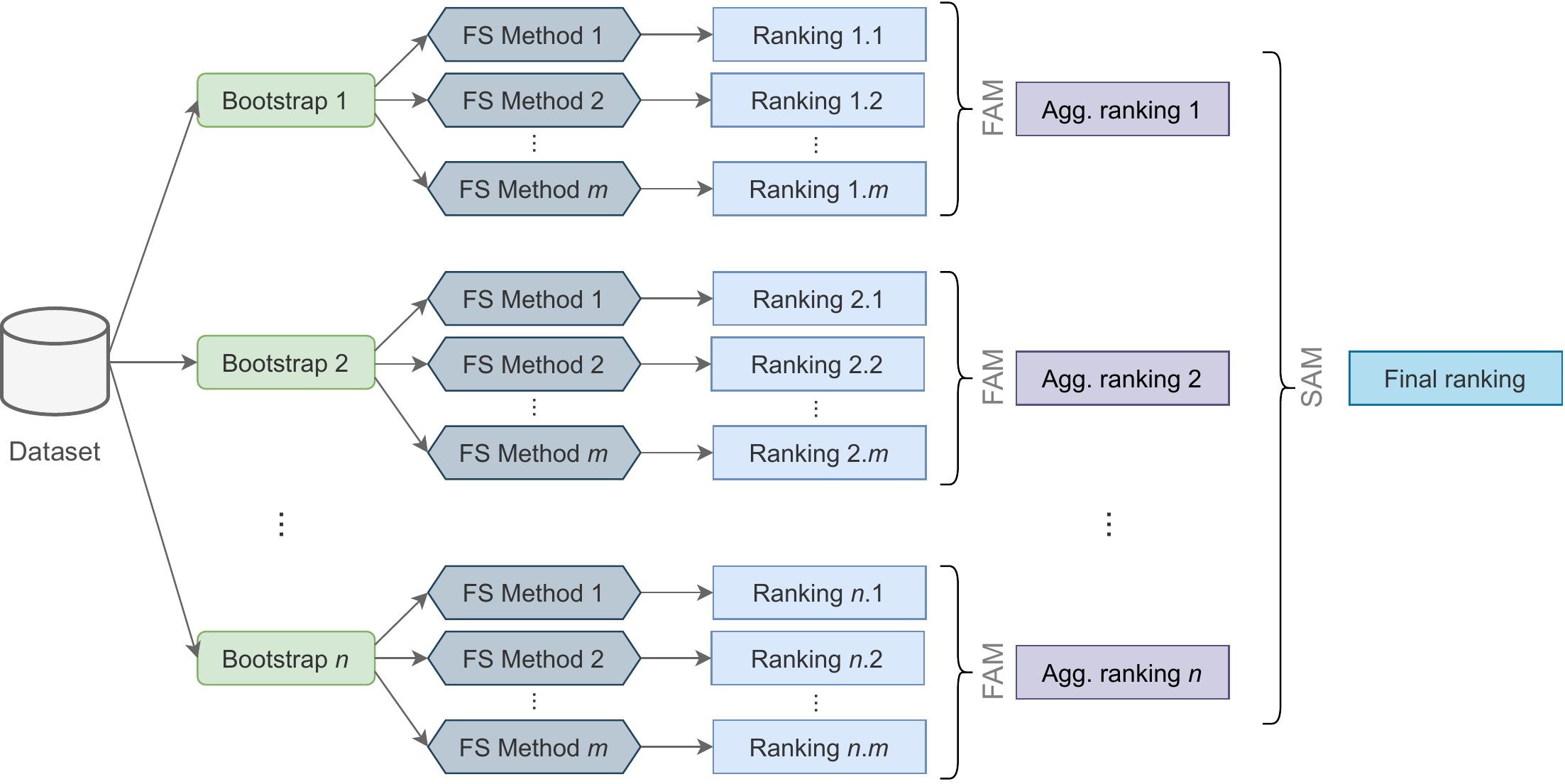}
    \caption{The Hybrid Ensemble Feature Selection design using \textit{m} FS methods exposed to \textit{n} bootstrap bags. Rankings aggregation is performed in a two-stage process: the first aggregation method (FAM) is applied within bootstraps, and the second aggregation method (SAM) is applied across bootstraps. FAM and SAM can be either the same or different aggregation strategies.}
    \label{fig:ens-hyb}
\end{figure*}

In our EFS proposal, we aim to accomplish diversity of feature relevance opinion in two ways: function perturbation and data perturbation. By mixing the heterogeneous and homogeneous approaches, we believe this new ensemble technique could explore even more diverse opinion and minimize any dependence in relation to a data sample, FS method, or a specific combination of both. Ultimately, our goal is to improve stability of selected features while keeping a good predictive potential. In whats follows we detail the proposed Hybrid Ensemble Feature Selection (Hyb-EFS) design, which is summarized in Figure~\ref{fig:ens-hyb}. Our approach implements the generation of diversity in two levels and performs ranking aggregation in two stages.

\subsection{Hybrid sources of diversity}

First, our Hyb-EFS design applies data perturbation over the original dataset.  This is achieved by the bootstrap resampling technique (\ie{} sampling with replacement). A total of $n$ bootstraps with same sample size as the original dataset are generated. Each bootstrap will serve as input data for FS methods. Here we consider $n=50$ bootstraps in light of some previous studies with homogeneous ensembles \citep{Pes2020,saeys2008robust,abeel2010robust}.  \citet{Pes2020} suggested that using 50 bootstraps for homogeneous EFS is a suitable choice considering the tradeoff between computational time and performance gain. We note, however, that $n$ is a configurable hyperparameter of our approach that could be tuned according to user needs.

After bootstraps are generated, our Hyb-EFS design applies function perturbation for each bootstrap sample. Function perturbation is established by simultaneous use of $m$ distinct FS methods. In this work, we chose $m=5$ different FS, as reviewed in Section~\ref{fs-methods}. Some of these FS methods are domain-general and have been widely used in previous FS and EFS studies (\ie{ }GR, SU, ReliefF), whereas others are new promising approaches specifically designed for large-scale gene expression data (\ie{ }Wx and GeoDE). 

The output of the second level of diversity injection is multiple (\ie{ }$n \times m$) complete rankings of the given dataset features according to their relevance for classification. Features are ordered from the most discriminative to the least ones. Nonetheless, the relative ordering may vary across rankings due to distinct training sets and to modifications in the FS method.


\subsection{Aggregation strategy}

After generating the multiple rankings, their aggregation must be carried out to draw a final decision about feature relevance. In this study, we propose to perform ranking aggregation in a two-stage process. The first aggregation derives a consensus ranking for each bootstrap sample in the ensemble. In other words, in the first stage we combine the results obtained upon applying five distinct FS methods to the same data sample. Since different selectors assume different criteria to rank features, a consolidated ranking among distinct methods may represent a better compromise of relevant features: if a feature is highly ranked by all selectors, it must be highly relevant to the problem. This aggregation stage reduces $n \times m$ rankings to a set of $n$ rankings.

In the second stage, the $n$ resulting rankings are merged into a single, final ranking that represents the Hyb-EFS's output. This aggregation aims at summarizing the findings for multiple samples of the data, given that many FS methods are sensitive to data variation. Therefore, after the second stage of aggregation, top ranked features are those consistently identified as relevant features for distinct samples of data and different FS methods. This approach is motivated by its good potential to improve stability of candidate biomarkers. 
We refer to the functions applied in the first and second stage of aggregation as First Aggregation Method (FAM) and Second Aggregation Method (SAM), respectively.

In this study, we compare two aggregation configurations for our Hyb-EFS method. The first configuration uses Borda Aggregation for both FAM and SAM. Borda Count is a popular voting rule that combines preferences of multiple voters and has been previously applied in the EFS context \citep{Drotar+2019}. The second configuration uses a Stability Weighted Aggregation as the FAM and the traditional Borda Aggregation as the SAM. In what follows we explain these functions in more details.

Assuming a ranking representation where the first position is taken by the most important feature, \ie the ranking has a descending order of relevance, an aggregation score takes into account those positions to calculate the final aggregation value for each feature. Let \(r\) be the number of feature rankings available to aggregate; \(N_f\) be the total number of features; \(p_{i,j}\) be the position of the feature \(i\) in the ranking \(j\); and \(AS_i\) be the aggregated relevance score for a particular feature \(i\). The $AS_i$ is computed for each feature and then a final relevance ranking, the aggregated one, is obtained by ordering the feature aggregated relevance scores.

We note that feature selection per se happens when a threshold \(th\) is applied to the generated rankings for selecting only the top \(th\) relevant features as the subset of the most explanatory ones. The two strategies adopted to compute the aggregated relevance scores are:
\begin{itemize}
    
    \item \textbf{Borda Aggregation} is based on the Borda Count voting system. It is computed by simply adding points to each feature $i$ inversely proportional to the position $j$ occupied in the ranking $r$ to generate their respective \(AS\).
    \[AS_i = \sum_{j=1}^{r} N_f - p_{i,j}\]
    
    \item \textbf{Stability Weighted Aggregation} is used only as FAM. Initially, given the desired threshold \(th\), it calculates the KI of the generated rankings across the bootstraps for each FS algorithm. Then, within each bootstrap, it aggregates the rankings originated from each FS algorithm by computing the score similar to the Borda count approach, further multiplying it by a transformed KI to the power of 5. This function was empirically defined to induce a FS algorithm's stability influence in the aggregation process in an exponential manner. Consequently, a slight increase in the stability makes a greater impact on the aggregation process than it would with a simple linear multiplication. The transformation process takes the KI interval \([-1,1]\) and maps it to \([0,2]\).
    Let \(KI_r\) be the transformed KI for a ranking \(r\), we can calculate \(AS_{i}\) in the Stability Weighted Aggregation method with the following equation:
    \[AS_{i} = \sum_{j=1}^{r} (N_f - p_{i,j}) \cdot KI_r^5\]
\end{itemize}

Because the proposed Stability Weighted Aggregation is only possible in the two-stage aggregation system of the Hyb-EFS, all other ensemble experiments reported in our work only used the Borda Aggregation as their aggregation method.

\section{Experiments}

In this section, we explain the experiments run to assess the potential of the proposed Hyb-EFS design. We detail all the EFS strategies used as baselines in our study as well as variants of the proposed approach, the evaluation scheme adopted, and the cancer datasets explored in the experiments.

\subsection{Ensemble feature selection strategies}

Our main goal with the proposed Hyb-EFS design is to be able to improve predictive potential and stability of candidate biomarkers drawn from large-scale data, such as transcriptome. Thus, it is important to compare our approach with previous strategies well-consolidated in the related literature. We considered four types of feature selection strategies during our experiments:

\begin{itemize}

\item \textbf{SingleFS}: this strategy refers to traditional FS methods applied without any type of ensemble configuration. Each of the algorithms listed in Section~\ref{fs-methods} is applied a single time over the training data to generate its ranking of feature relevance. These results are referred to as Sin-\textit{X}, where \textit{X} is a FS algorithm described in Section~\ref{fs-methods}.  So, for example, Sin-SU refers to the results of the SU algorithm.

\item \textbf{Hom-EFS}: each one of the considered FS algorithms is also used in their homogeneous ensemble version built with a fixed number of bootstraps (we adopted $n = 50$). Since we are using five algorithms as base selectors in our approach, we generate five distinct Hom-EFS strategies. We refer to Hom-\textit{X} as the Hom-EFS implementation of the FS algorithm \textit{X}. So, for example, Hom-SU is a Homogeneous Ensemble Feature Selection using Symmetrical Uncertainty as the base selector.

\item \textbf{Het-EFS}: this strategy builds a heterogeneous EFS approach using all the considered FS algorithms as base selectors in its setup. No bootstraps are used in this case, since diversity is introduced by function perturbation. Thus, the input for each FS method is the complete training data. We also tested a variant of the Het-EFS stratgy that uses only Wx, GR, and SU as base selectors, that we refer to as \textbf{Het Wx-GR-SU}. These methods were chosen after empirical verification of their good performance.

\item \textbf{Hyb-EFS}: these experiments refer to the proposed strategy of Hybrid EFS, in which data and function perturbation are jointly explored. Data perturbation is achieved by drawing $n= 50$ bootstraps from the training dataset, whereas function perturbation is carried out by using all the FS methods listed in Section~\ref{fs-methods} as base selectors. Different from the previous strategies, the Hyb-EFS involves a two-stage aggregation process. Here, we implement and compare two variants of the proposed Hyb-EFS design: \textbf{Hyb-EFS-Borda} adopts the Borda Aggregation both as FAM and SAM; whereas \textbf{Hyb-EFS-Stb} uses the Stability Weighted Aggregation as FAM and the Borda Aggregation as SAM. We also implement a variant of the hybrid approach that uses only Wx, GR, and SU algorithms for the function perturbation process (\textbf{Hyb Wx-GR-SU}) to allow a direct and fair comparison with the its corresponding heterogeneous EFS strategy (Het Wx-GR-SU).

\end{itemize}

\subsection{Evaluation scheme}

All the FS strategies listed in the previous section were compared under the same methodology to assess predictive potential and stability. To avoid the ``peeking phenomenon", which refers to data contamination that occurs when data used for testing the model is seen during some part of the training process, we carefully followed the proper protocol for FS proposed by \citet{kuncheva2018feature}. This protocol suggests that FS should be carried out inside the cross-validation loop. The main purpose of adopting this protocol is to avoid reporting optimistically biased estimate of the classification performance. However, it also reduces the chance of selecting feature subsets that are artefacts rather than truly discriminative features.

Therefore, in our framework, the FS process runs under a stratified $k$-fold cross-validation process that provides new unseen data to evaluate models built with features selected by one FS strategy. We adopted \textit{k = 5} following previous recommendation \citep{james2013introduction}. This number of folds also considers that some datasets may have low-sample-size for some classes, a common characteristic in omics data, which would result in low representation of the minority class in each fold and thus a poor performance evaluation for this specific class if more folds are used.

\begin{figure*}[!ht]
\centering
    \includegraphics[width=0.87\textwidth]{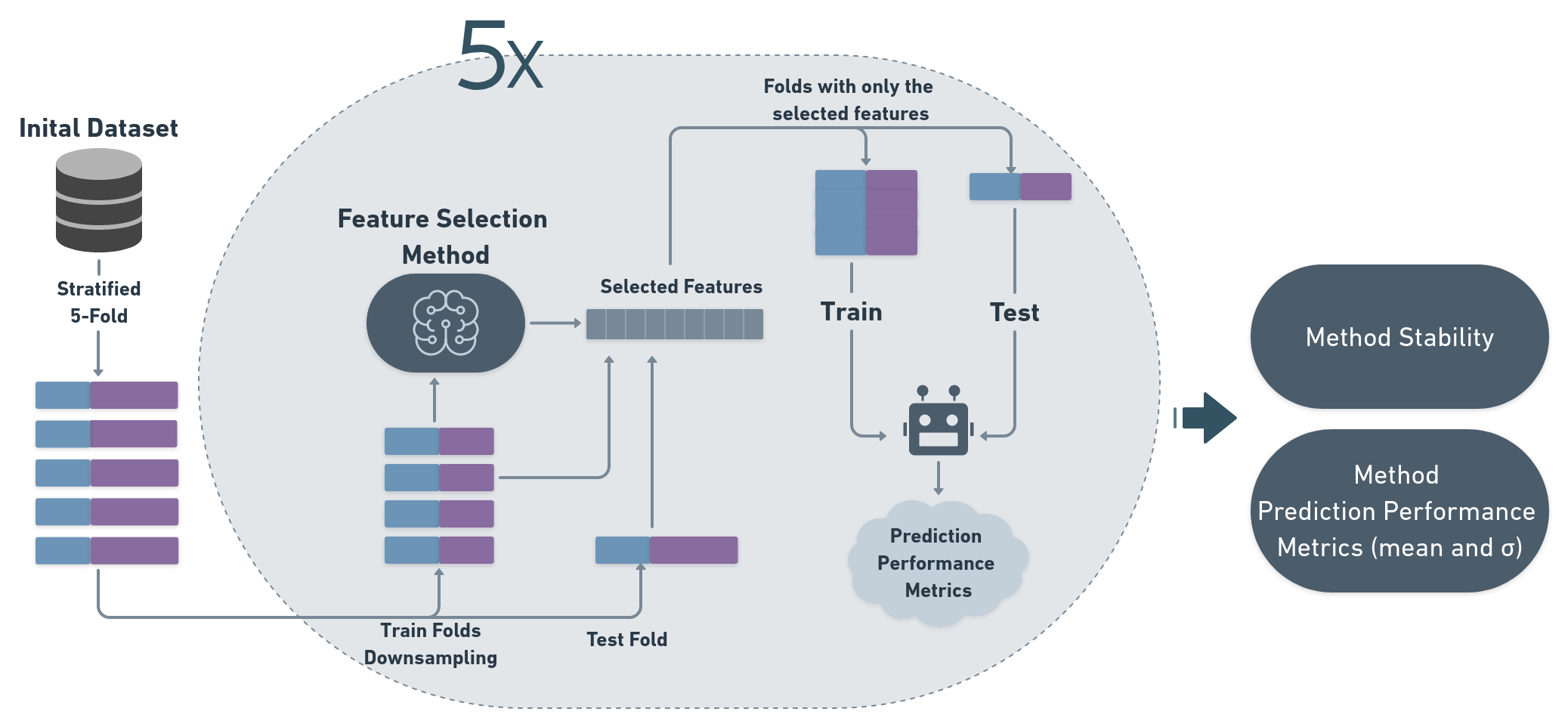}
    \caption{The evaluation scheme implementing a proper feature selection protocol inside a process of stratified cross-validation with downsampling.}
    \label{fig:scv-process}
\end{figure*}

Our evaluation scheme is summarized in Figure~\ref{fig:scv-process}. Inside the cross-validation loop, after the division of the original dataset in $k$ folds, class imbalance is corrected by applying downsampling to the train folds. Each fold is preprocessed to have a 50:50 balanced data by randomly removing samples from the majority class. After that, the train folds are combined into a training set and used as input for the FS process.

Each FS strategy outputs a final ranking of features in decreasing order of relevance according to its specific criterion. Features are selected by applying a threshold \(th\) to this ranking, \ie{ } only the top \(th\) relevant features are retained in the further steps of the cross-validation iteration. Our experiments varied \(th\) in the interval from 1 to 50 extensively. Nonetheless, we also considered the top 75, 100, 200, and 500 features (shown in the supplementary material). For each subset of selected features, we assessed its stability with the KI metric. Additionally, its predictive potential was evaluated by the classification performance obtained with a Gradient Boosting Machine (GBM) classifier trained only with the subset of selected features. The GBM classifier is an ensemble method that builds additive and sequential regression trees. New models are fitted to improve the estimate of the target value based on the errors of previous models \citep{Friedman2001}. Instead of identifying prediction errors made by prior models using adaptive weights for dataset instances, as in Adaboost, GBM uses the gradient of the loss function, which is a measure of how good is the model to fit the underlying data. We note that any classification algorithm could be used at this step; GBM was chosen because it is among the top methods for supervised binary classification problems.

By evaluating the classifier's performance with the test fold, any and multiple performance metrics can be extracted. Here, we adopted the ROC AUC and PR AUC scores. Nonetheless, other metrics could also be considered together in order to obtain a broader understanding of the FS method's predictive potential. Each iteration of the $k$-fold cross-validation process returns an estimate of the performance metric, and at the end results are summarized as the mean and standard deviation across the $k$ iterations.


\subsection{Datasets}

In this study, we are especially interested in investigating biomarkers related to cancer. A large volume of transcriptome data has been made publicly available by the scientific community in databases such as ArrayExpress \citep{10.1093/nar/gky964}. To evaluate our approach and contribute to new hypotheses regarding candidate cancer biomarkers, we obtained merged microarray-acquired datasets generated by \citet{Lim+2019} for  breast (E-MTAB-6703), liver (E-MTAB-6695), lung (E-MTAB-6699), and pancreas (E-MTAB-6690) cancer. Meta-datasets were compiled by the authors integrating several independent studies for each type of cancer, all of them using the Affymetrix Human Genome U133 Plus 2.0 Array to quantify gene expression levels. 

The main characteristics of the datasets used in our experiments are outlined in the Table \ref{datasets-summary}. Their large sample-size serves as an excellent  ``discovery cohort" for research towards new candidates for diagnostic biomarkers. Each meta-dataset has been previously pre-processed by authors using standard bioinformatics pipeline as described in their paper \citep{Lim+2019}. Data distribution indicated normalized characteristics, presenting aligned medians and interquartile ranges (Figure S1). Therefore, no further pre-processing step was required.

\begin{table}[!ht]
\centering
\resizebox{0.6\columnwidth}{!}{

\begin{tabular}{@{}lllll@{}}
\toprule
                    & \textbf{Breast} & \textbf{Liver} & \textbf{Lung} & \textbf{Pancreas} \\ \midrule
\#Features          & 20545           & 22881          & 20545         & 22881             \\
\#Tumor samples     & 2088            & 264            & 1474          & 108               \\
\#Non-Tumor samples & 214             & 137            & 147           & 70                \\
\#Total samples     & 2302            & 401            & 1621          & 178               \\ \bottomrule
\end{tabular}
}
\caption{Summary of the selected datasets used in our experiments. Features relate to genes, and samples relate to the instances of our dataset.}
\label{datasets-summary}
\end{table}

\section{Results}
\label{results}

In the following subsections we discuss the performance of the proposed hybrid EFS approach, comparing it against each of the counterpart strategies (\ie, SingleFS, Hom-FS, and Het-FS). Since we observed a better performance in terms of predictive potential and stability of our Hyb-EFS-Stb in contrast to the Hyb-EFS-Borda, our discussion will be focused on this variant of the hybrid EFS approach.

We also note that despite extracting a set of predictive potential metrics from all strategies, the analysis will mostly discuss the methods' ROC AUC, to allow a qualitative comparison with previous works, as well as their PR AUC (plots are provided in the supplementary material). 
We anticipate that due to the reduced sample size of collected metrics (\ie five for each experiment configuration based on a 5-fold cross-validation), statistical tests were impracticable for spotting robust statistically significant differences between the experiments' performance.

Based on a general analysis of the overall results across all experiments, we noticed that the ROC AUC and other performance metrics were quite similar among distinct strategies in a large body of comparisons. This prevented us from surely elicit the best method based on visual inspection of PR AUC distributions in some cases. Thus, the analysis will be mainly oriented towards the most stable method, which in most cases can be visually distinguishable, making it possible to select the best method (or methods) based on the plots. All results were evaluated based on the top \(th\) ranked genes, with \(th\) varying in the interval [1,50]. A broader analysis considering thresholds in the interval [25,500] is provided in the supplementary material.

 

Figure \ref{fig:res-distcv} depicts the behavior of the predictive potential in terms of ROC AUC and PR AUC for all the experiments reported in this study. We observed that the liver and pancreas datasets were the most challenging to classify. In addition, the ReliefF FS method, both in the SingleFS and in its Hom-EFS version, had the worst classification performance across all datasets compared to all other methods. Looking closely, we also noticed that the Sin-GeoDE and the Hom-GeoDE methods have a slightly lower median in the challenging datasets in contrast to other methods, except for ReliefF. The proposed approaches, Hyb-EFS-Borda and Hyb-EFS-Stb (higlighted in blue), showed good consistency in performance and were competitive with other EFS approaches despite having the caveat of presenting more outliers for the liver and pancreas datasets compared with, for instance, Hom-SU and Hom-GR.

\begin{figure*}[h!]
   \centering
   \subfloat[ROC AUC]{\includegraphics[width=0.48\textwidth]{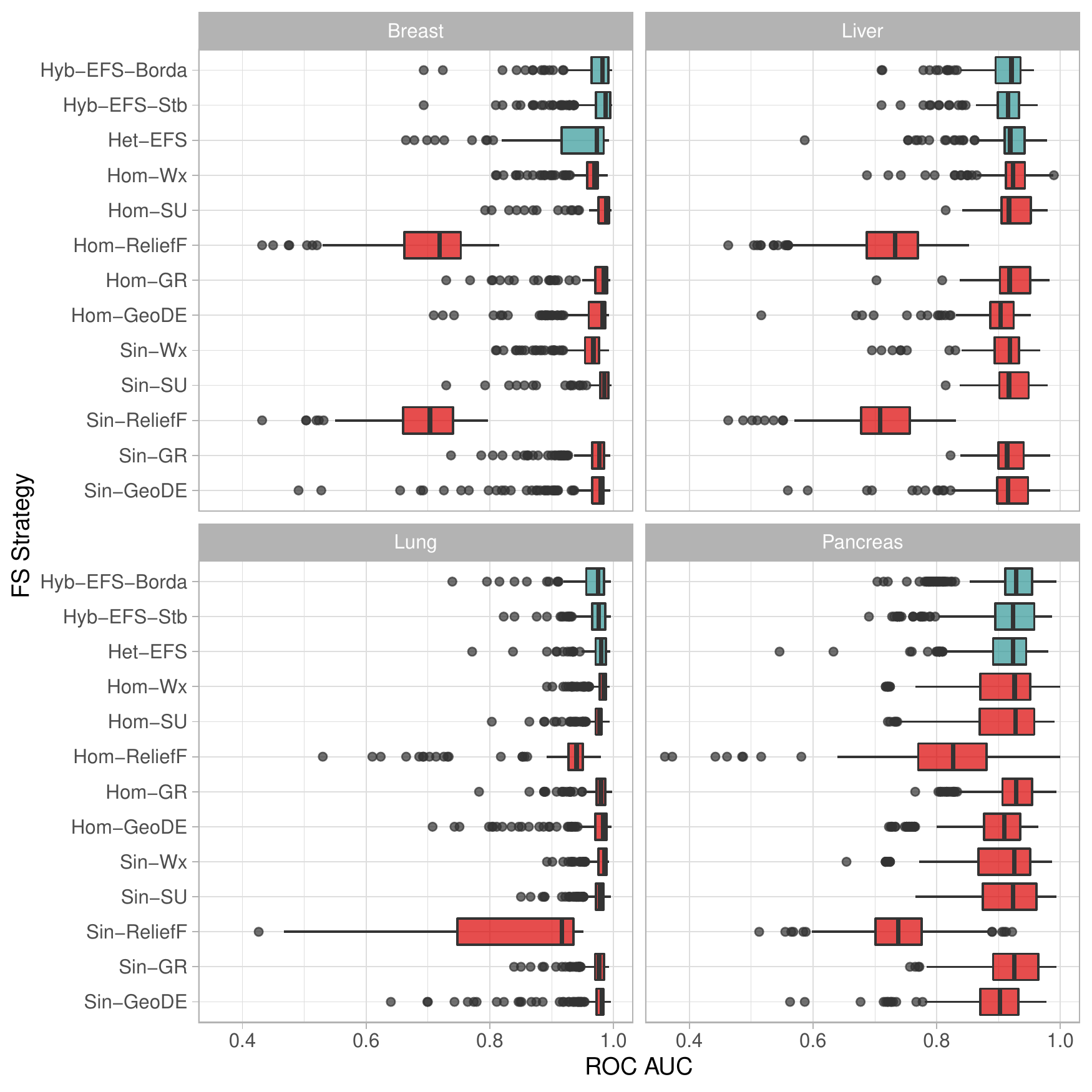}}
   \label{fig:res-distcv-rocauc}
   \quad
   \subfloat[PR AUC]{\includegraphics[width=0.48\textwidth]{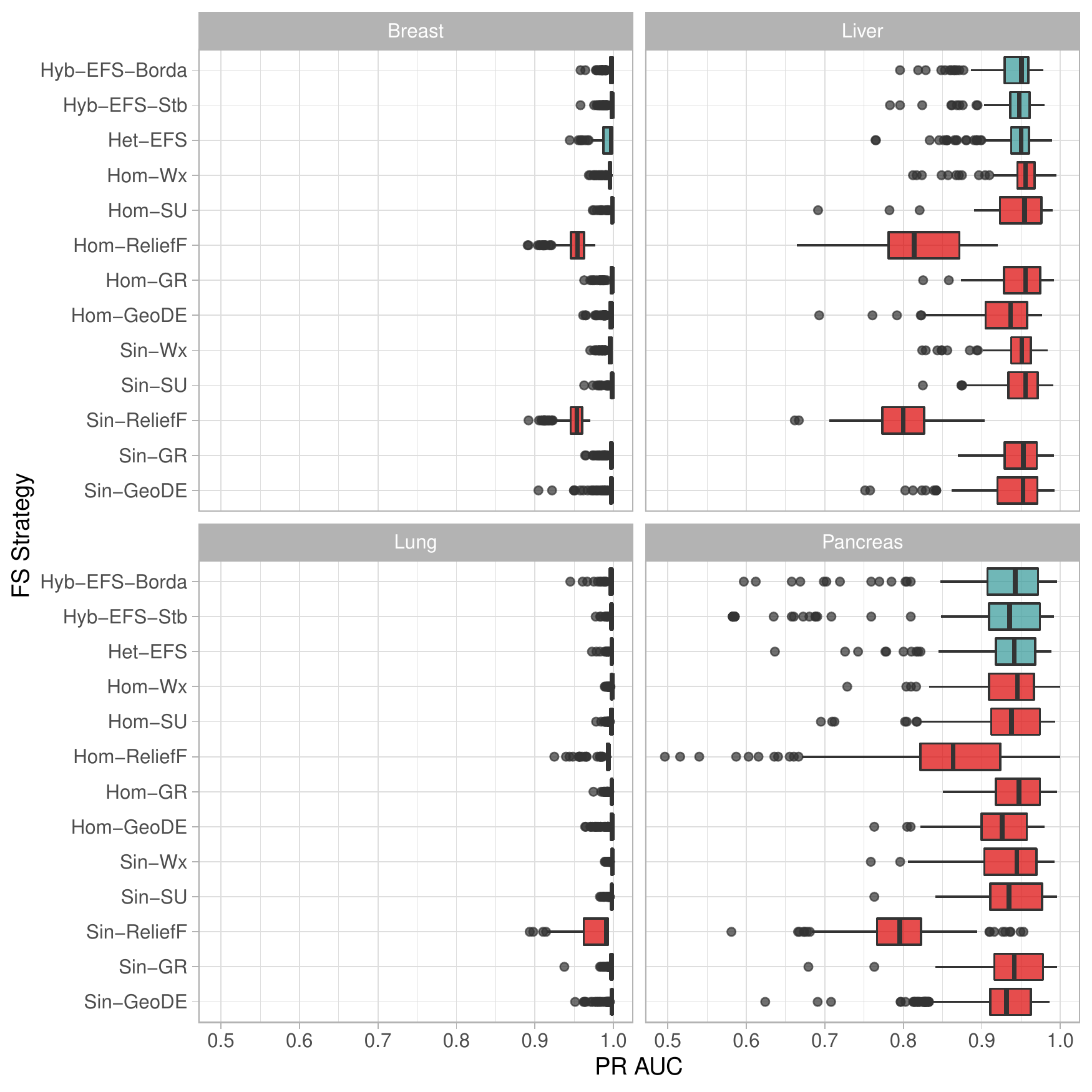}
   \label{fig:res-distcv-prauc}}
    \caption{Distribution of ROC AUC and PR AUC score across five folds, considering all thresholds $th \in [1,50]$. The hybrid EFS variants proposed in the current work are highlighted in blue.}
\label{fig:res-distcv}
\end{figure*}

\subsection{Hyb-EFS vs SingleFS}
\label{subsec:hyb-singlefs}

In comparing Hyb-EFS and the base FS methods, it is imperative to keep in mind that any ensemble-based solution will be more computationally expensive than single FS methods - and our hybrid EFS approach is not an exception for this. This may call into question whether EFS approaches would really payoff the downside of higher computational costs. The decision of whether single or ensemble FS approaches will be used is certainly influenced by the availability of computational resources. Nonetheless, comparing their performances is relevant in the scientific context to understand how the individual components of our approach behave and how robust is the Hyb-EFS method to any deficiency presented by single FS.

\begin{figure*}[h!]
   \centering
   \subfloat[Stability]{\includegraphics[width=0.48\textwidth]{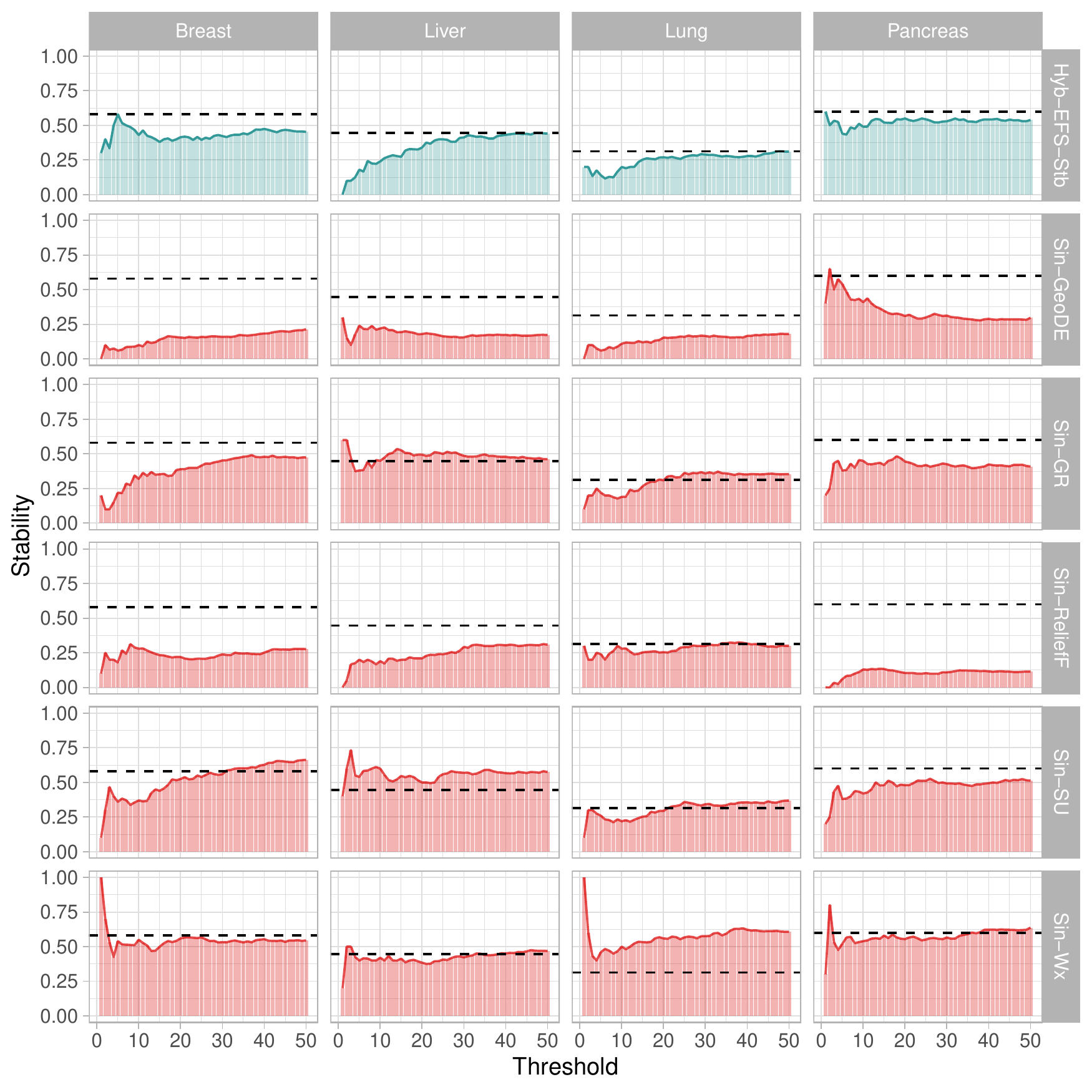}
   \label{fig:res-stb-hyb_vs_single}}
   \quad
   \subfloat[ROC AUC]{\includegraphics[width=0.48\textwidth]{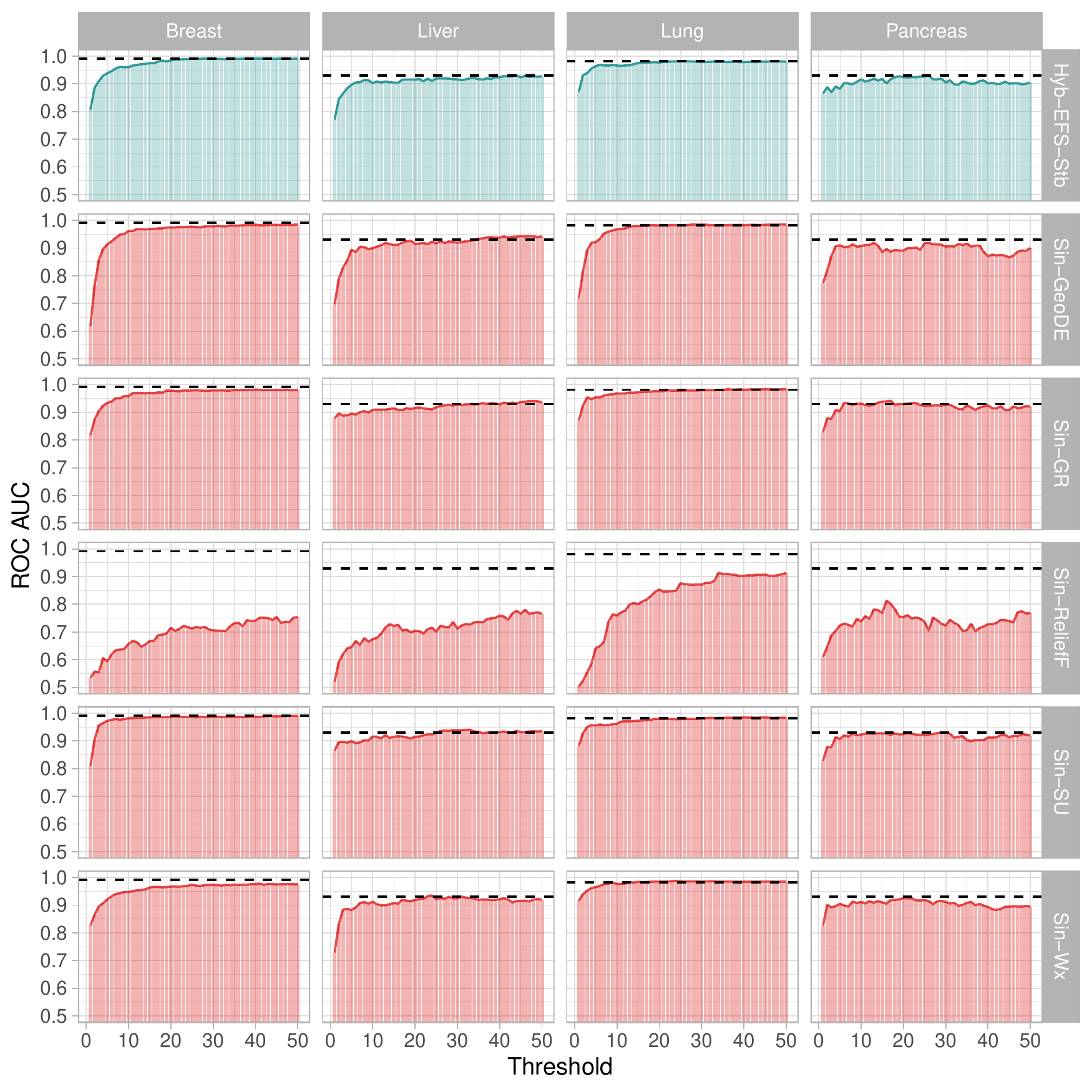}
   \label{fig:res-rocauc-hyb_vs_single}}
    \caption{Hyb-EFS-Stb and SingleFS comparison in terms of KI's stability and the mean ROC AUC computed over the 5-fold cross-validation process. The proposed approach is highlighted in blue. The dotted line represents the maximum value achieved by the Hyb-EFS-Stb in each dataset for each metric. It is used as a reference point to compare the proposed approach to other FS methods.}
\label{fig:res-hyb_vs_single}
\end{figure*}

In Figure \ref{fig:res-hyb_vs_single}, Hyb-EFS-Stb and SingleFS methods are compared in terms of a) stability and b) ROC AUC across all datasets. The dotted line denotes Hyb-EFS-Stb's maximum values achieved for each dataset, allowing us to identify if its performance is surpassed by the counterpart methods and, if so, whether this occurs towards the top of the ranking (denoted by lower thresholds).

There is no clear best method for the ROC AUC performance, but some conclusions may be drawn. We observed, again, that the Sin-ReliefF presented the worst performance. Sin-GeoDE and Sin-Wx also had some flaws for the classification of pancreas cancer cases, with ROC AUC values below the best values achieved for Hyb-EFS-Stb for most thresholds analyzed. Nonetheless, we reinforce that this is a more challenging dataset. Even our hybrid approach had some deterioration in the ROC AUC for this thresholds interval in the pancreas dataset. An interesting thing to note is that despite the significantly lower ROC AUC of the ReliefF algorithm, the Hyb-EFS-Stb approach was robust enough to deal with its poor performance and keep a good ROC AUC score. The Hyb-EFS-Stb almost entirely corrected the mistakes induced by the ReliefF. A similar effect was observed for the PR AUC score, although with more variation in Hyb-EFS-Stb's outcome (Figure S2).

Regarding the stability of each method, the hybrid ensemble excelled Sin-ReliefF and Sin-GeoDE approaches. The other SingleFS experiments based on the GR, SU, and Wx algorithms were better for the liver dataset. Moreover, the Sin-Wx was, in general, the best single FS method, showing huge improvement for the lung dataset. Although this is true for the four datasets used here, we highlight the findings of \citep{seijo2017ensemble} showing that, across multiple different domains, there was no SingleFS robust enough to keep a performance as good on average as their heterogeneous ensemble. Thus, there is no guarantee that the Wx will replicate this performance on other datasets. However,according to previous works, this could be the case for the ensembles taking advantage of the function perturbation approach (hybrid and heterogeneous ensembles) as their performance tend to be more stable across distinct domains.

\subsection{Hyb-EFS vs Hom-EFS} 
\label{subsec:hyb-homefs}

\begin{figure*}[h!]
   \centering
   \subfloat[Stability]{\includegraphics[width=0.48\textwidth]{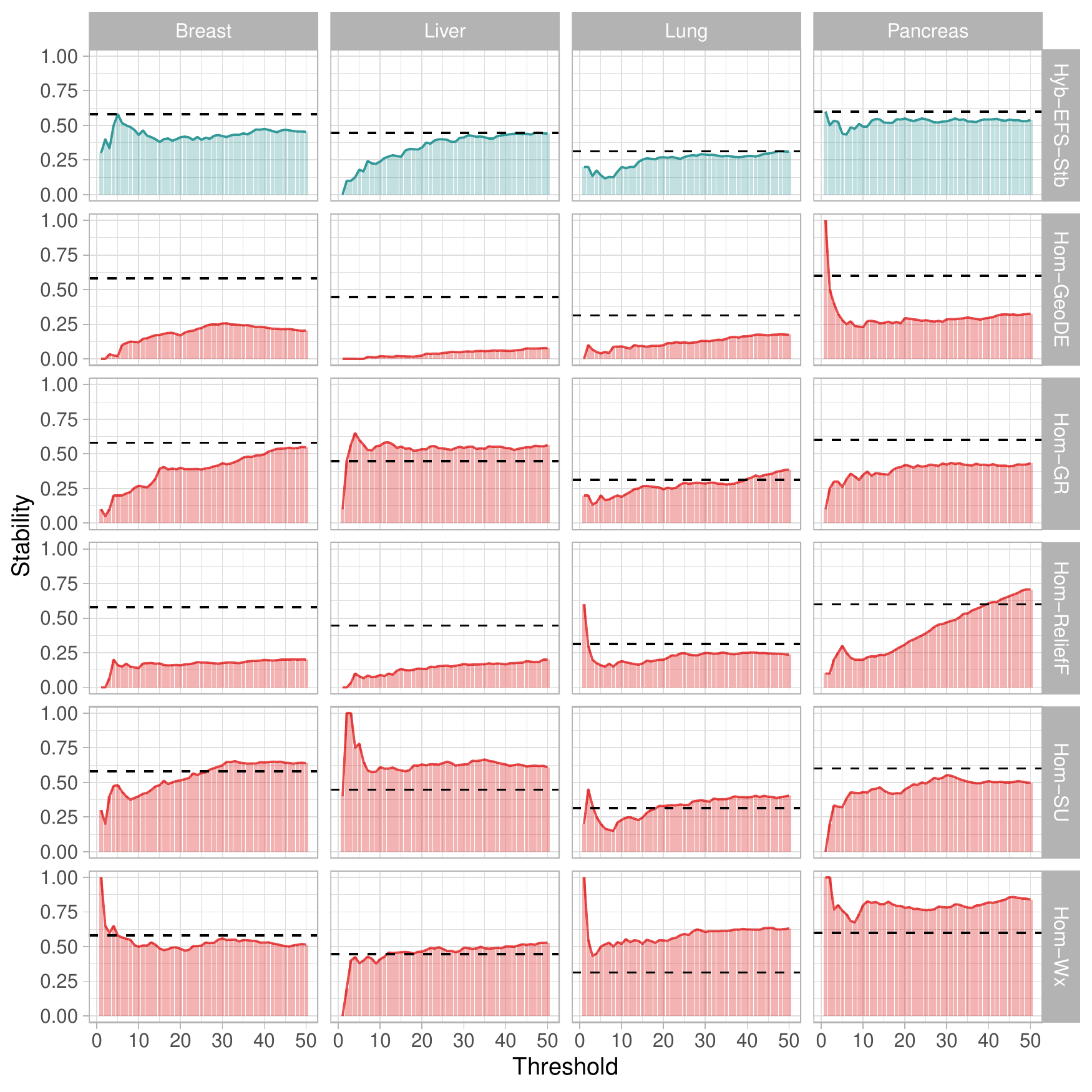}
   \label{fig:res-stb-hyb_vs_hom}}
   \quad
   \subfloat[ROC AUC]{\includegraphics[width=0.48\textwidth]{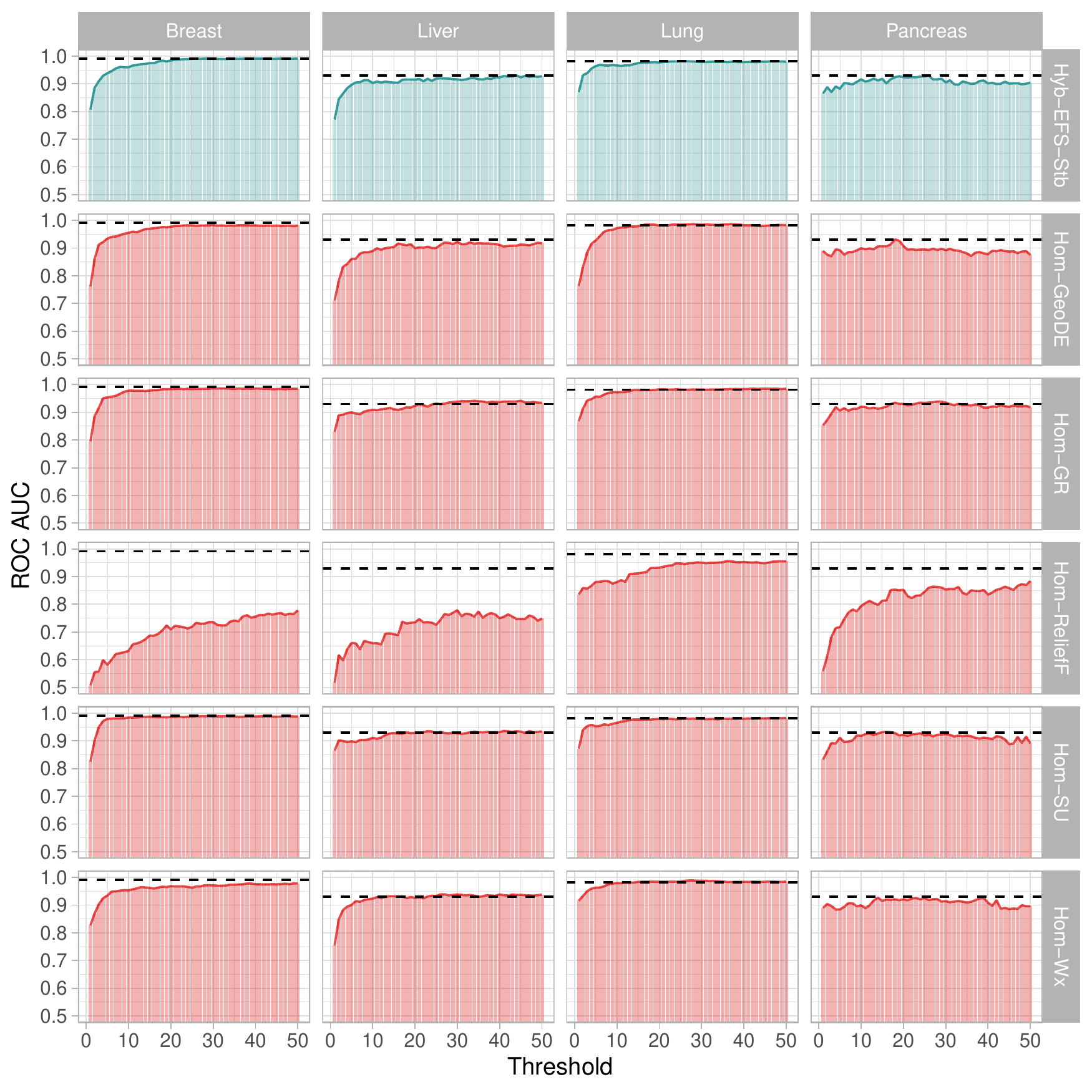}
   \label{fig:res-rocauc-hyb_vs_hom}}
    \caption{Hyb-EFS-Stb and Hom-EFS comparison in terms of KI's stability and the mean ROC AUC computed over the 5-fold cross-validation process. The proposed approach is highlighted in blue. The dotted line represents the maximum value achieved by the Hyb-EFS-Stb in each dataset for each metric. It is used as a reference point to compare the proposed approach to other FS methods.}
\label{fig:res-hyb_vs_hom}
\end{figure*}

The comparison between the hybrid and homogeneous EFS designs yielded similar insights in relation to the previous SingleFS comparison. As shown in Figure \ref{fig:res-hyb_vs_hom}, the Hom-Wx is the best overall method, with consistent ROC AUC score and stability across all datasets. Its stability was superior than our Hyb-EFS-Stb approach, especially in the lung and pancreas datasets. Interestingly, its stability was boosted in relation to its SingleFS version (\ie Sin-Wx) for the pancreas scenario, where it clearly outperforms the Hyb-EFS-Stb. We note that a greater stability for homogeneous EFS methods may be expected to some extent since our approach has two sources of diversity, which potentially increases the variation among the multiple features relevance ranking, thus impacting on stability. 

We observed that the GeoDE and the ReliefF algorithms also benefited from the homogeneous ensemble for the pancreas cancer dataset, particularly in relation to stability (Figure \ref{fig:res-stb-hyb_vs_hom}) and PR AUC (Figure S3). Nonetheless, for the liver dataset, these two algorithms lost stability, and even the Hom-Wx was negatively impacted in this scenario when we closely analyze the first thresholds applied (\ie [1,5], see Figure S3). The Hom-SU also deserves some credit, since it was the most stable method for the liver cancer data, also showing a good performance for both ROC AUC and PR AUC scores. 

However, the bigger picture revealed that the FS methods without the function perturbation suffer more from stability variations across datasets, in contrast to what is observed for our hybrid EFS approach. Even under poor results presented by its base FS algorithms, the Hyb-EFS-Stb was robust enough to keep good stability and predictive potencial in different data scenarios, as discussed in Section~\ref{subsec:hyb-singlefs}.

\subsection{Hyb-EFS vs Het-EFS} 
\label{subsec:hyb-vs-het}

\begin{figure*}[h!]
   \centering
   \subfloat[Stability]{\includegraphics[width=0.48\textwidth]{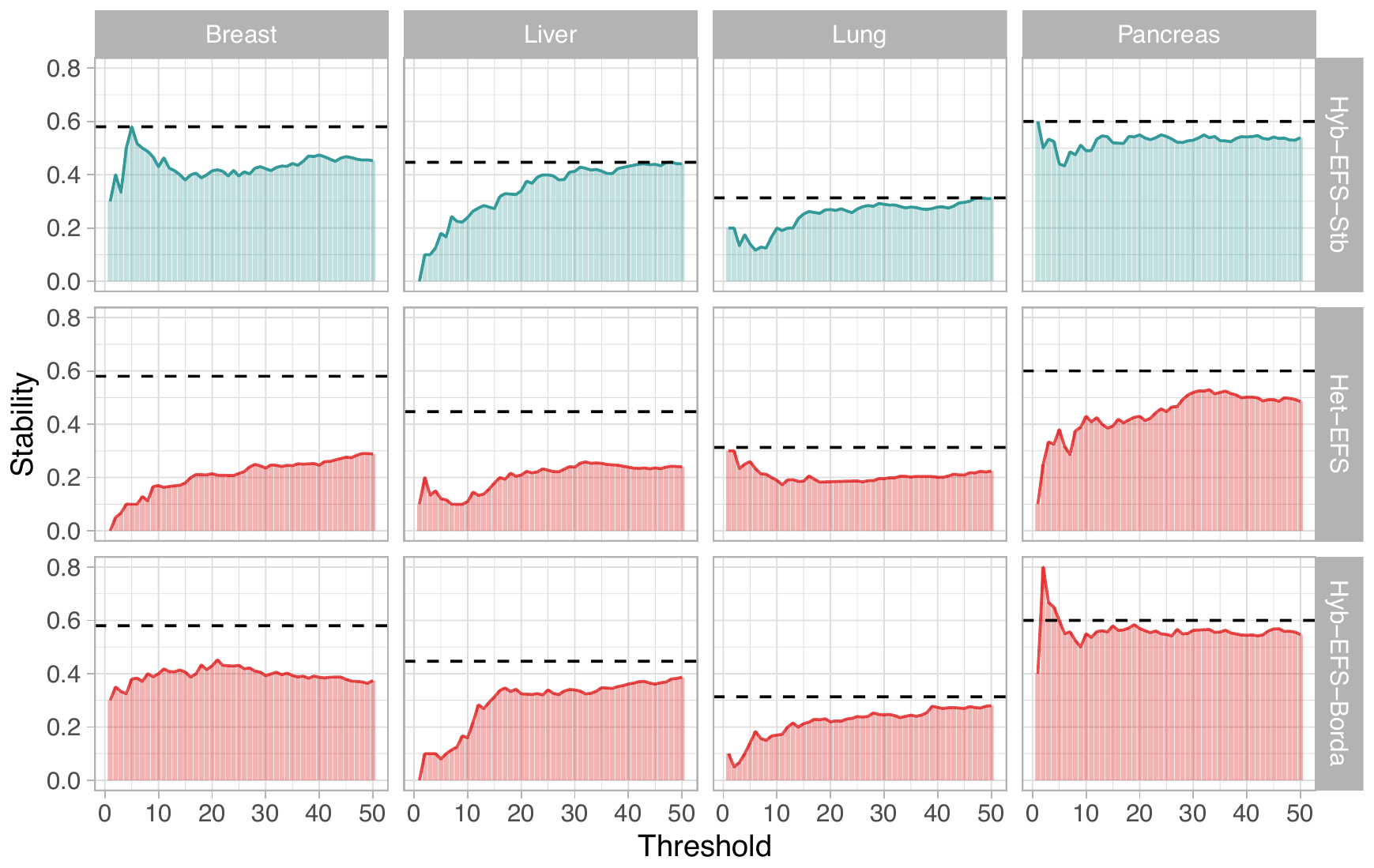}
   \label{fig:res-stb-hyb_vs_het}}
   \quad
   \subfloat[ROC AUC]{\includegraphics[width=0.48\textwidth]{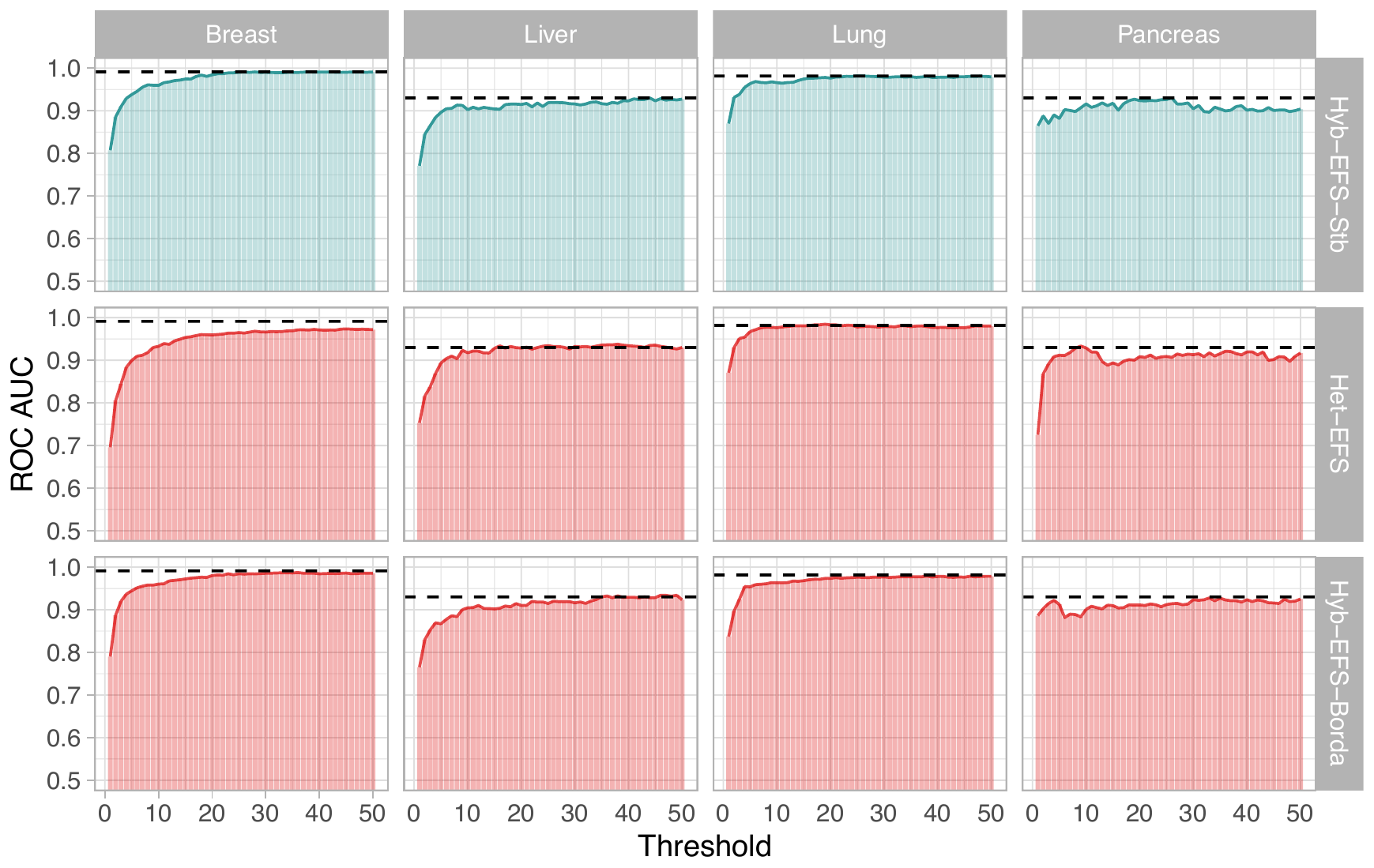}
   \label{fig:res-rocauc-hyb_vs_het}}
    \caption{Hyb-EFS and Het-EFS comparison in terms of KI's stability and the mean ROC AUC computed over the 5-fold cross-validation process. The proposed approach is highlighted in blue. The dotted line represents the maximum value achieved by the Hyb-EFS-Stb in each dataset for each metric. It is used as a reference point to compare the proposed approach to other FS methods. We also show the performance for the Hyb-EFS-Borda approach.}
\label{fig:res-hyb_vs_het}
\end{figure*}

Figure \ref{fig:res-hyb_vs_het} compares the two hybrid ensemble approaches to the heterogeneous ensemble. When analyzing stability, we observed a remarkable difference, with a clear advantage for both hybrid EFS methods in contrast to the Het-EFS. Except for the initial thresholds of the lung cancer dataset, all other scenarios pointed to the Hyb-EFS as the most stable method. Regarding which hybrid approach is the best one, only in the pancreas cancer data the Hyb-EFS-Borda seemed to improve the performance over the Hyb-EFS-Stb for stability, ROC AUC, and PR AUC (Figure S4).

The main reason why researchers should choose the Het-EFS strategy is to free themselves from having to select the best method for their specific domain problem, \ie{}, having to experiment and evaluate different methods. Despite the poor performance of some FS algorithms for the data at hand, others could provide great opinions regarding the features' relevance, thus justifying the usage of a combined opinion obtained from a Het-EFS. Nonetheless, the experiments performed here clearly indicated the lack of stability a Het-EFS could deliver in some scenarios like ours. The hybrid ensemble approach breaks through this limitation by highly increasing the Het-EFS stability while maintaining or even improving its predictive potential.

This is actually a very important finding. The heterogeneous ensemble design is known to be the most reliable method for FS across multiple domain data \citep{ali2018uefs, dittman2012comparing, seijo2017ensemble}, keeping comparable results with the best Single FS methods in most cases. Hence, in face of our findings, we believe that the proposed Hyb-EFS design improves over the Het-EFS method, pushing forward the state-of-the-art in FS strategies that can free researchers and other data science professionals from having to experiment and select the most appropriate FS algorithm for their target data, which may be cumbersome and time-consuming.

\subsection{Hyb-Wx-GR-SU vs Het-Wx-GR-SU vs Hom-Wx} 

\begin{figure*}[h!]
   \centering
   \subfloat[Stability]{\includegraphics[width=0.48\textwidth]{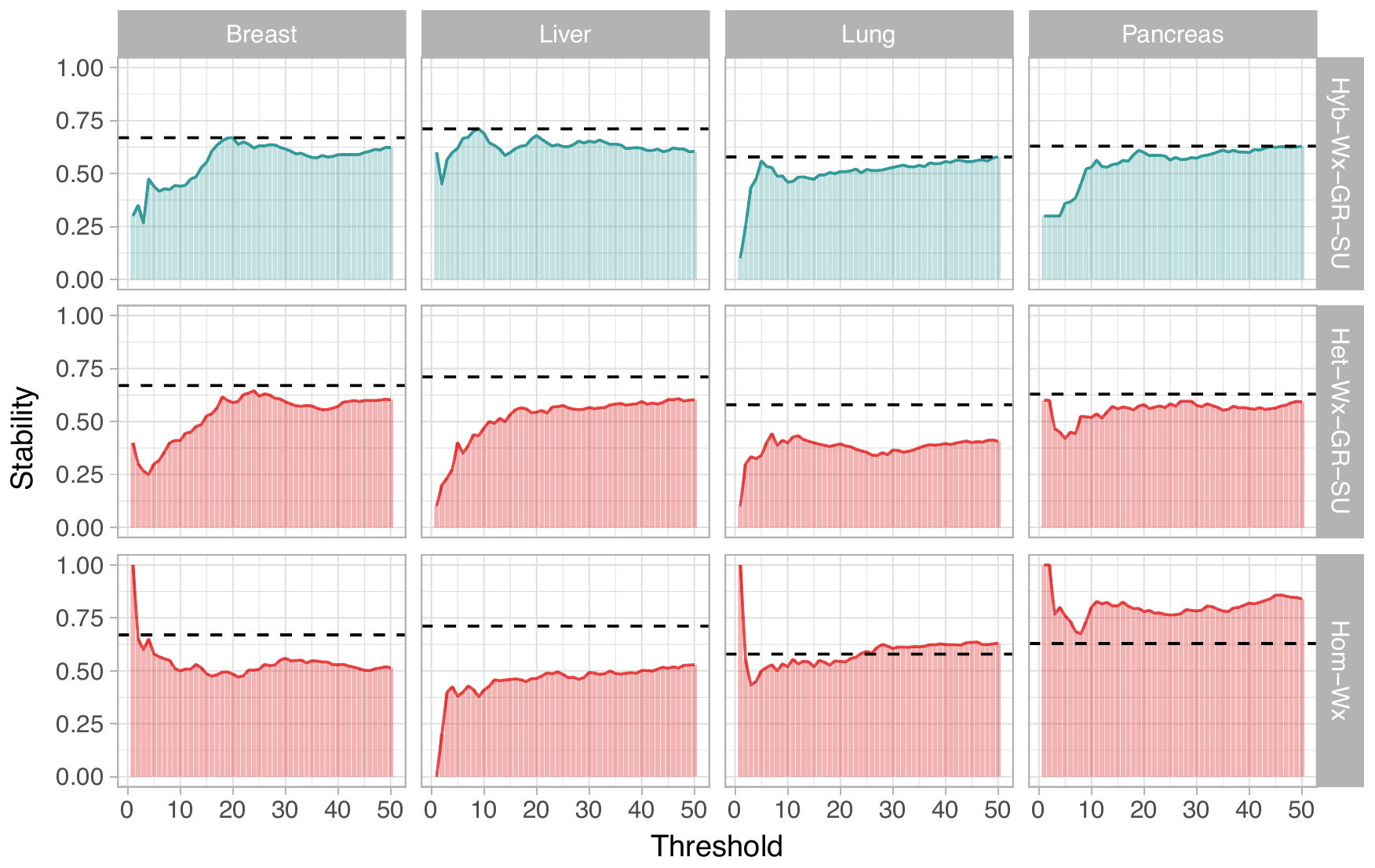}
   \label{fig:res-stb-hyb_vs_het3}}
   \quad
   \subfloat[ROC AUC]{\includegraphics[width=0.48\textwidth]{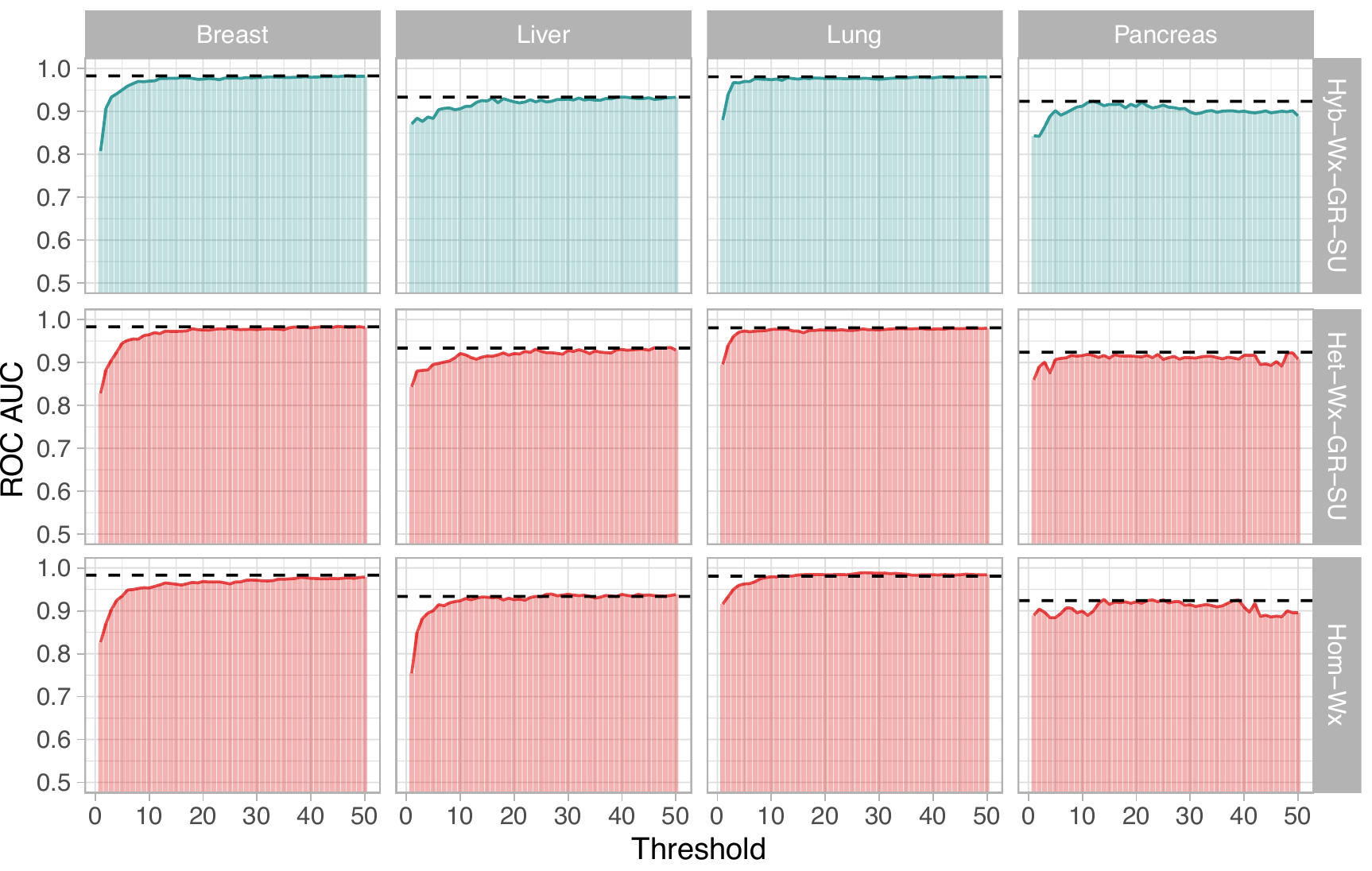}
   \label{fig:res-rocauc-hyb_vs_het3}}
    \caption{Hyb-Wx-GR-SU, Het-Wx-GR-SU and Hom-Wx methods performance comparison in terms of KI's stability and the mean ROC AUC considering the five folds of the cross-validation process. The dotted line represents the maximum value achieved by the Hyb-Wx-GR-SU in each dataset for each metric and it is used for comparing the performances among the other FS approaches.}
\label{fig:res-hyb_vs_het3}
\end{figure*}

In our previous experiments, the ReliefF and GeoDE methods seemed to be significantly worse in the selected datasets both in the SingleFS and Hom-FS experiments. This poor performance certainly impacts the Hyb-EFS-Stb and Het-EFS approaches, despite observing certain robustness to ineffective methods in our Hyb-EFS-Stb method. Thus, we tested variations of the hybrid and heterogeneous approaches that do not use these methods as base selectors, which we refer to as Hyb-Wx-GR-SU and Het-Wx-GR-SU, respectively. We compared these designs to the Hom-Wx, which was the most consistent method for stability and predictive potential across the four cancer datasets. Figure \ref{fig:res-hyb_vs_het3} shows the results for this comparison.

Regarding stability, we noticed notorious improvements brought by the Hyb-Wx-GR-SU. It achieved similar stability as the Het-WX-GR-SU in the breast cancer data but superior stability in the liver and lung datasets. Moreover, Hyb-Wx-GR-SU was less stable for lower thresholds but more stable towards higher thresholds in the pancreas dataset. Except for a few initial thresholds, the Hyb-Wx-GR-SU was also notably better than the Hom-Wx for the breast and liver datasets. In the lung dataset, Hom-Wx achieved the highest stability among all three approaches for lower thresholds. However, results were comparable for threshold five onwards. Finally, Hom-Wx was particularly well-suited for the pancreas cancer data and achieved the best stability across all methods and thresholds. Of note, the pancreas cancer dataset was a challenging one and for this scenario we considered the Hom-Wx as the most stable method among all the experiments performed.

In terms of predictive potential, few discrepancies were found among the three methods compared. In the liver dataset, Hyb-Wx-GR-SU and Het-Wx-GR-SU were better than Hom-Wx for the initial thresholds (\ie in the interval [1,5]), whereas for the pancreas dataset Hyb-Wx-GR-SU was slightly inferior than the other two approaches. This was observed both for ROC AUC and PR AUC scores (Figure S5). We also noticed that the Hyb-Wx-GR-SU method fixed most of the PR AUC low outliers faced by Hyb-EFS-Stb in the pancreas dataset (Figure S4). These results suggested that an improvement of the Hyb-EFS approach to allow an automatic selection of base FS methods during features relevance analysis could enhance its overall performance even more.

\subsection{The big picture}

The analysis of our results provided distinct insights in each closer look. In some cases, Wx and SU were the best methods, albeit, in others, the Hyb-EFS showed improved performance over these base FS algorithms. The best overall method seemed to be the Hom-Wx. However, when the two worst FS algorithms were removed from the heterogeneous and hybrid ensembles, there was a significant positive impact on the outcomes. The hybrid approach seemed to surpass the Hom-Wx both in the breast and liver cancer datasets.

As mentioned in Section~\ref{subsec:hyb-vs-het}, the heterogeneous ensemble was previously regarded by the literature as the best approach for FS across many different datasets and domains. Among other reasons, such as its outstanding performance, this is related to the lack of necessity of choosing a specific FS method without a success guarantee in the domain of interest or having to compare multiple FS methods experimentally, which is costly. Thus, we emphasize that there is no guarantee that the Hom-Wx will always show the good stability and predictive potential observed in our work. Here we provided evidence that the hybrid design improves the heterogeneous approach in different datasets. Additionally, we outlined that the two-level aggregation system of the hybrid ensemble design could be explored to automatically select the best methods among the bootstraps and aggregate only their opinions, resulting in, for example, a Hyb-Wx-GR-SU with no 
manual interference needed. This could be a promising direction for future research.

\begin{figure*}[!h]
   \centering
   \subfloat[Stability]{\includegraphics[width=0.98\textwidth]{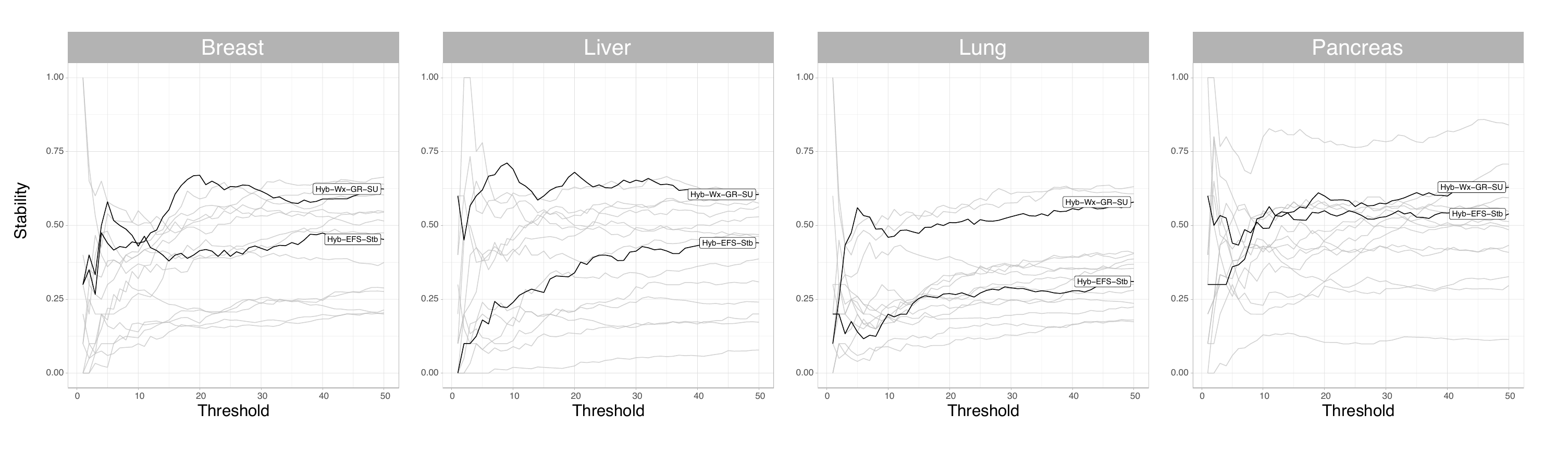}
   \label{fig:res-stb-hyb_vs_all}}
   
   \subfloat[ROC AUC]{\includegraphics[width=0.98\textwidth]{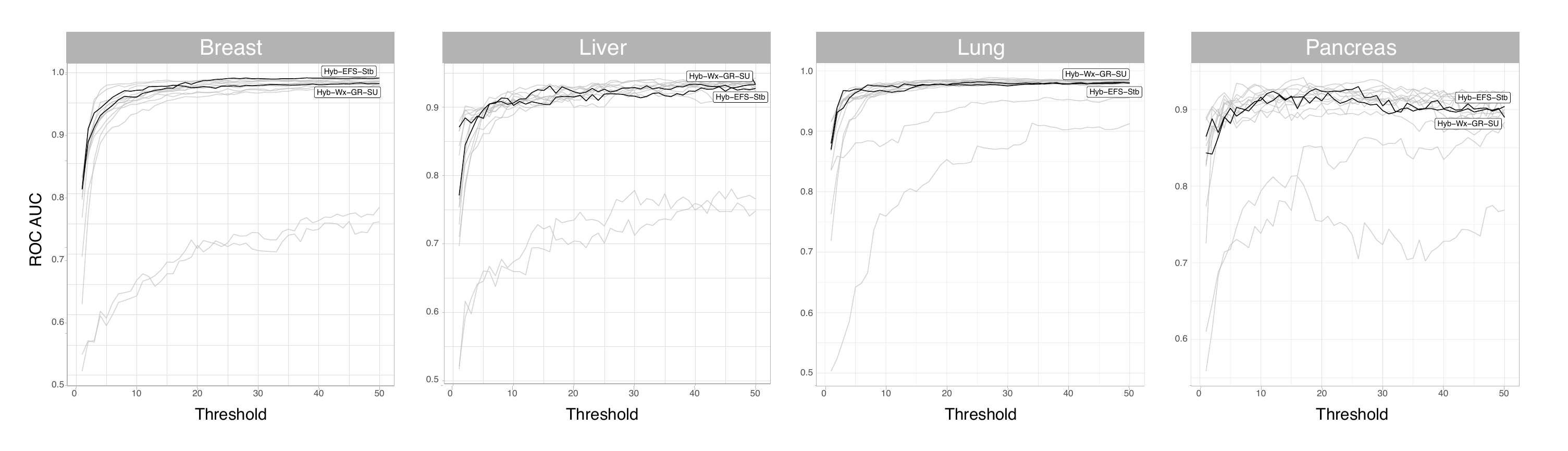}
   \label{fig:res-rocauc-hyb_vs_all}}
    \caption{Hyb-EFS-Stb and Hyb-Wx-GR-SU (outlined) compared to all the other FS approaches considered in the present study.}
\label{fig:res-hyb_vs_all}
\end{figure*}

Our results and observations can be summarized by Figure \ref{fig:res-hyb_vs_all}, which highlights the stability and ROC AUC of both the Hyb-EFS-Stb and the Hyb-Wx-GR-SU approaches against all experimented methods. We can see how successful the hybrid approach with the three top methods was compared to the other ones, keeping up in the top stabilities for most of the thresholds among 1 to 50.

\section{Discussion of the biological plausibility of findings} 
\label{sec:biologicalPlausibility}

Since our paper has a particular interest in pointing candidate biomarkers for cancer diagnosis from microarray data, to further evaluate our findings and contribute to the literature of biomarker discovery methods, we investigated if the reported genes align with what is already well-documented for the explored types of cancer. We adopted bioinformatics databases and tools to verify if cancer-related genes and pathways are enriched in the ranking produced by the proposed Hyb-Wx-GR-SU approach. We chose this hybrid EFS method due to its outstanding performance for both criteria adopted, namely stability and predictive potential. 

Enrichment analyses were carried out with the Gene Set Enrichment Analysis (GSEA) method using the clusterProfiler R package \cite{Wu+2021}. In summary, GSEA is a validated method to obtain signature genes related to a disease by considering a ranking over genes evaluated \cite{Subramanian+2005}. Gene sets are defined based on prior biological knowledge, \eg annotated databases or literature, and are compared against a given ranking to determine whether their members tend to occur towards the ranking's top (or bottom). Here, we are specifically interested in finding enrichment of gene sets at the top of our ranking, which indicates that the EFS method correctly prioritizes cancer-related genes. The estimated significance level provided by GSEA is adjusted for multiple hypothesis testing using the False Discovery Rate (FDR) method. An FDR $\leq$ 0.05 was considered statistically significant in all subsequent analyses. For comparison purposes, we also run GSEA analysis for the other two top-performing ensemble methods, Het-Wx-GR-SU and Hom-Wx.

\begin{figure*}[!h]
  \centering
  \subfloat[NCG]{\includegraphics[width=0.3\textwidth]{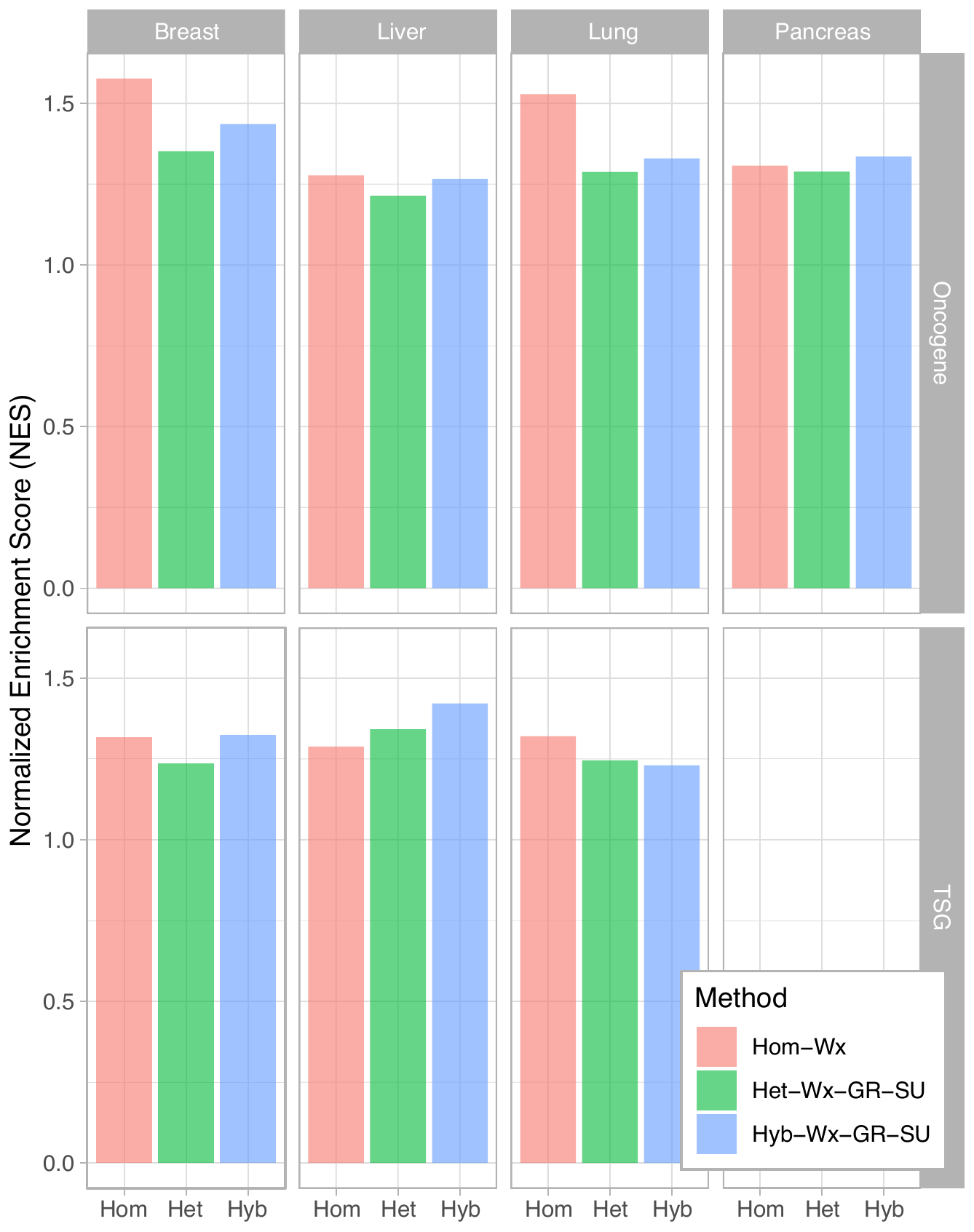}
  \label{fig:res-bioinfo-ncg6}}
  \quad
  \subfloat[MSigDB-C6]{\includegraphics[width=0.3\textwidth]{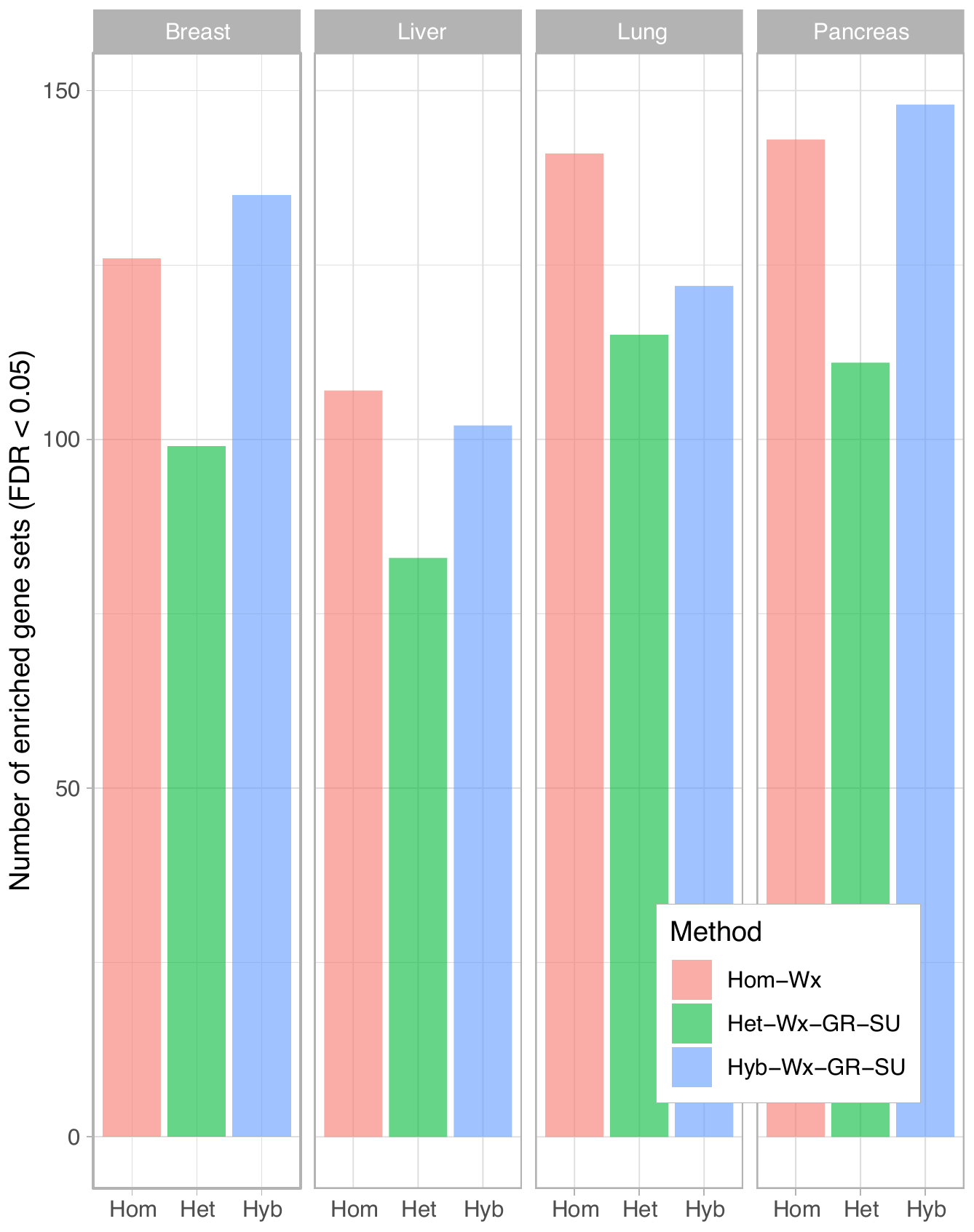}
  \label{fig:res-bioinfo-c6}}
  \quad
  \subfloat[KEGG]{\includegraphics[width=0.3\textwidth]{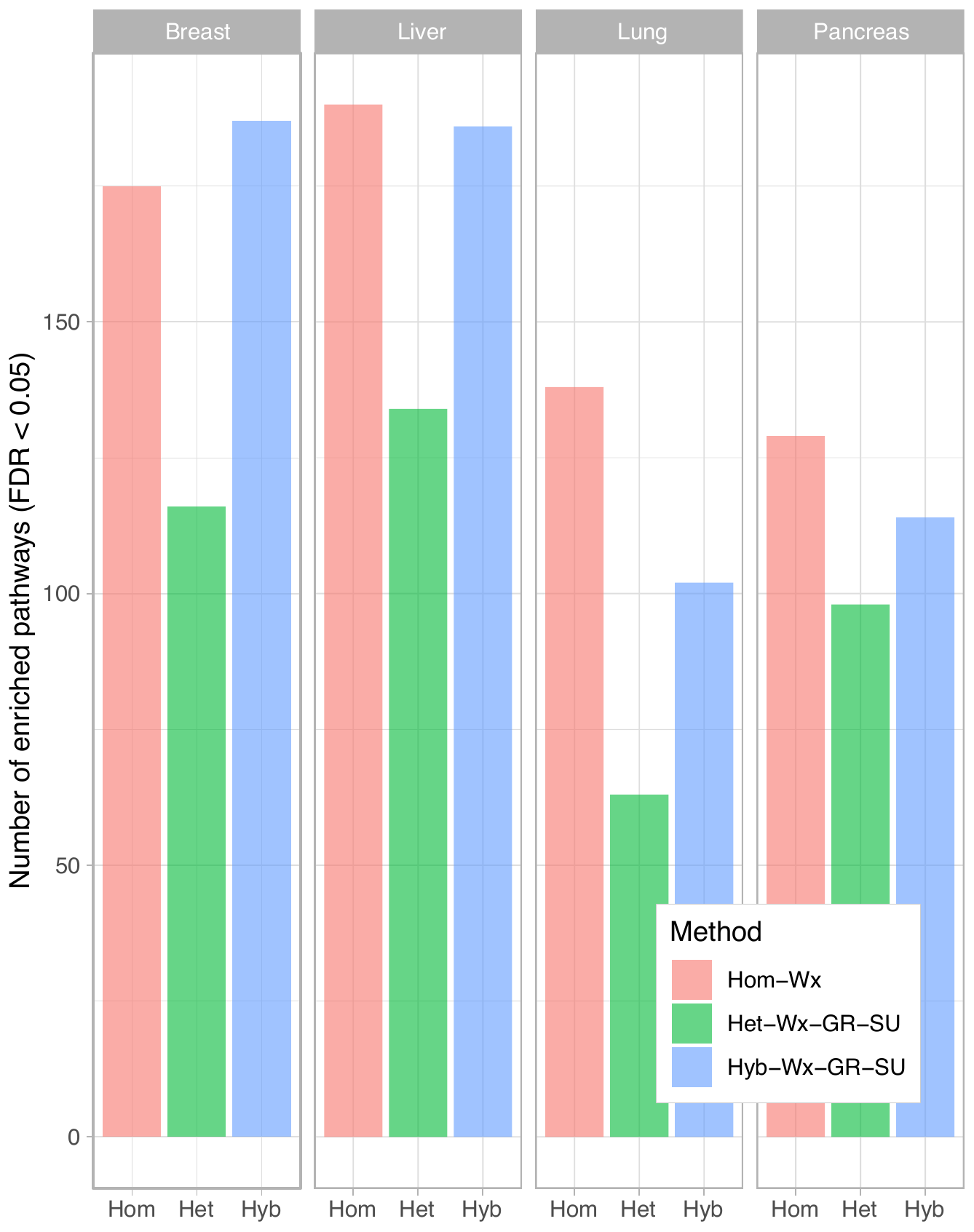}
  \label{fig:res-bioinfo-kegg}}
    \caption{Comparing functional enrichment analysis of rankings obtained with the top-three performing FS methods using a) driver genes from NCG, b) cancer genes signature from MSigDB, and c) pathways annotations from KEGG.}
\label{fig:res-bioinfo}
\end{figure*}

Cancer cells are characterized by intense proliferation and reduced programmed cell-death. This lack of control occurs because of mutations in genes that modulate cell growth and division, named cancer driver genes. There are two classes of cancer driver genes: oncogenes and tumor suppressor genes (TSG). Oncogenes are responsible for cell proliferation, hence mutations in these genes result in more cell division than necessary. TSG have the opposite role, being responsible for impeding excessive cell division; when TSG are mutated the cells lose their ability to control their own proliferation, resulting in tumorigenesis \cite{Hanahan&Weinberg2000,Hanahan&Weinberg2011,Vogelstein+2013}. According to the Network of Cancer Genes (NCG) database \cite{Repana+2019}, there are 711 genes confirmed as Oncogenes or TSG. 
Thus, we run GSEA to investigate if both gene sets, Oncogene or TSG, were enriched for each of the tumors evaluated, using the top-performing methods (Figure \ref{fig:res-bioinfo-ncg6}).

For the three methods, we were able to enrich the Oncogenes in the four tumors studied considering an FDR $\leq$ 0.05. Except for TSG in pancreas, for which none of the methods compared achieved significant results, all other methods and datasets showed statistically significant enrichment for TSG. We concentrate our comparative analysis on the Normalized Enrichment Score (NES), which reflects the degree to which a gene set (oncogenes or TSG) is overrepresented at the extreme of the ranking - here, towards the top of the ranking. The Hom-Wx method presented the highest NES for oncogenes enrichment in breast and lung, as well as TSG enrichment in lung, with clear advantage over the other two methods. Our Hyb-Wx-GR-SU approach was very competitive against Hom-Wx in the other scenarios. Moreover, Hyb-Wx-GR-SU performed slightly better than the Het-Wx-GR-SU method in all scenarios except for TSG in lung cancer. 

There was no enrichment of TSG in pancreas tumor considering the methods here investigated (FDR $>$ 0.05), which may be influenced to some extent by the smaller number of samples available for this type of cancer. However, the absence of enrichment might limit the application of these methods in pancreatic cancer, especially considering \textit{p53} is a TSG. \textit{P53} is the most studied cancer gene, and has an important role in pancreas tumorigenesis and tumor aggressiveness \citep{makohon2016pancreatic}. These biological validations suggest that the type of cancer evaluated and the sample size are relevant factors that may influence the outcome of the FS methods for biomarker discovery.

Also aiming to comprehend the biological relevance of our method, we evaluated a database that is considered a molecular signature of cancer, proposed by the developers of the GSEA \cite{Subramanian+2005} and available in the Molecular Signature Database (MSigDB). This database, identified as C6 collection, has 189 gene sets representing signatures of cellular pathways that are often dis-regulated in cancer. Considering the number of enriched gene sets, the Hyb-Wx-GR-SU method also performed better than Het-Wx-GR-SU for the MSigDB-C6 collection (Figure \ref{fig:res-bioinfo-c6}). Moreover, Hyb-Wx-GR-SU enriched a higher number of gene sets than Hom-Wx for breast and pancreas cancer datasets. The greater number of significantly enriched gene sets for Hom-Wx and Hyb-Wx-GR-SU as compared to Het-Wx-GR-SU is notable for all cancer datasets.

Finally, we conducted another GSEA to evaluate the biological role of genes that are not known as cancer drivers, but may be involved in important processes related to tumorigenesis. To accomplish this, we used a list of genes and their biological signaling pathways obtained in the Kyoto Encyclopedia of Genes and Genomes (KEGG) database \cite{Kanehisa&Goto2000}. When considering the significant associations (FDR $\leq$ 0.05), Hyb-Wx-GR-SU method enriched more pathways than Het-Wx-GR-SU in all cancer types (Figure \ref{fig:res-bioinfo-kegg}), with a significant difference observed for breast, liver, and lung. For breast cancer, Hyb-Wx-GR-SU was the most enriched method among the top-three methods. Moreover, we observed that in general, the difference between the Hom-Wx and the Hyb-Wx-GR-SU method is much smaller than both of them compared to Het-Wx-GR-SU.

\begin{figure*}[!h]
  \centering
  \includegraphics[width=\textwidth]{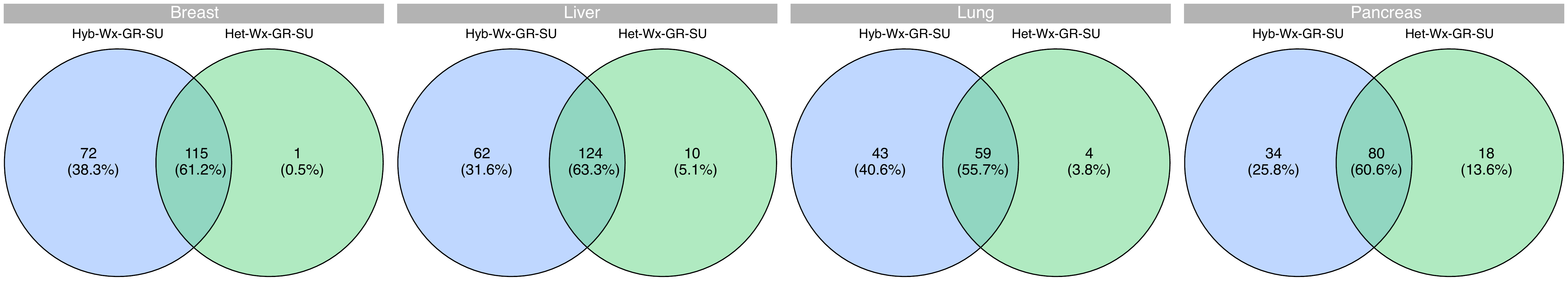}
  \caption{Venn diagrams comparing the number of enriched pathways in the GSEA for the Hyb-Wx-GR-SU and the Het-Wx-GR-SU methods.}
  \label{fig:res-bioinfo-venn}
\end{figure*}

To better understand these pathways and the functional differences of the top genes ranked by the two EFS approaches that explore function perturbation, we compared the enriched terms between Hyb-Wx-GR-SU and Het-Wx-GR-SU methods through Venn Diagrams (Figure~\ref{fig:res-bioinfo-venn}). The results were even more positive for the proposed method, indicating a greater plausibility in its findings when compared to its heterogeneous opponent, which is currently considered the most reliable FS strategy across multiple domains \cite{seijo2017ensemble}. First, the majority of the pathways enriched by the Het-Wx-GR-SU method were also enriched in our Hyb-Wx-GR-SU method; however, the opposite is not true. Second, the Hyb-Wx-GR-SU method performed better enriching cancer-related pathways than the Het-Wx-GR-SU method, such as the pathways of p53 \citep{kaur2018role} in breast cancer, mTOR \citep{ferrin2020activation}  in liver cancer, JAK-STAT \citep{pastuszak2017immunoexpression} in lung cancer, and HIF-1 \citep{hao2015hif} in pancreas cancer (see Supplementary Material). 

Only in the pancreas cancer dataset, the Hyb-Wx-GR-SU missed some important pathways for this type of tumor, \ie the cGMP-PKG signaling pathway. However, the identification of early diagnosis or precision treatment biomarkers in pancreatic tumor is considered a challenge when compared to other tumors, including breast, lung, colon, and cervix. It is believed that the asymptomatic nature of the tumor, until a very advanced stage, is one of the explanations for the lack of better screening \citep{hasan2019advances}. Hence, we conclude that the reduced performance of the Hyb-Wx-GR-SU method in pancreatic cancer can be explained when considering inherent difficulties related to the tumor etiology and its smaller sample size. 

An overall analysis of GSEA results suggests that the Hom-Wx method, which was the most stable across our experiments, achieved the highest biological significance of its findings across the considered gene sets (\ie NCG, MSigDB - C6, and KEGG). This observation corroborates the common sense that a high FS stability is indeed necessary to potentialize the utility of FS methods in biomarkers discovery. Nonetheless, as previously mentioned, homogeneous methods inherit from their base FS algorithms the uncertainty as to the superiority in performance across different domains. Thus, there is no guarantee that Hom-Wx will maintain the best performance for other data types or even other omics-related problems. In this sense, the heterogeneous method is our main baseline, given its potential to mitigate the generalizations shortcoming of the homogeneous-based approach, as highlighted in recent literature \cite{seijo2017ensemble}. The advantage of Hyb-Wx-GR-SU over its opponent Het-Wx-GR-SU was clear both for the number of enriched terms from MSigDB (C6) and KEGG databases and for the capacity of recovering important cancer-related pathways.

\section{BioSelector and efs-assembler tools}
\label{tools}
In addition to the comprehensive comparison among several feature selection strategies, including the proposed hybrid approaches, our work aimed to contribute to bioinformatics and machine learning fields by providing software resources to be used in future works with similar aims. We developed a highly flexible framework for performing experiments akin to ours with all the setups used in this research. The framework integrates Python (the primary programming language) and R to achieve higher flexibility and expressive power.

The framework comprises the \texttt{efs-assembler}\footnote{\url{https://github.com/colombelli/efs-assembler}}, a Python package implementing various functionalities for EFS analysis, and \texttt{BioSelector}\footnote{\url{https://github.com/colombelli/bioselector}}, a graphical user interface solution written using Electron/React (widely used desktop front-end solutions in industry) to support multiple operating systems and an user-friendly experience. Among the configurable parameters, users can choose basic configurations such as the ensemble design, the base FS algorithm(s), the datasets to extract the features, and the aggregation strategy. Users can also set parameters such as the number of bags for the bootstrap resampling (if applicable), the seed for reproducibility purposes, the classifier model to evaluate the rankings, the threshold values to be applied for feature selection, the number of folds in the cross-validation, and whether downsampling should be applied to balance classes. 

The \texttt{efs-assembler} package (executed as part of \texttt{BioSelector}'s backend) offers up to six FS algorithms and two classifiers provided by different Python and R packages. Gain Ratio and Symmetrical Uncertainty are implemented by the FSelectorRcpp R package \citep{fselectorrcpp}; Characteristic Direction was offered in MATLAB, Mathematica, R, and Python (the version used in our framework) by the same authors who proposed the method \citep{clark2014characteristic}; Wx was made available in a Python script by its authors \citep{park2019wx}, but some code modifications were necessary to execute it properly in our environment; ReliefF is a Python package hosted in PyPI repository\footnote{https://pypi.org/project/ReliefF/}, written by the authors of scikit-rebate \citep{Urbanowicz2017Benchmarking}, which integrates the ReliefF algorithm with other FS methods\footnote{Due to runtime problems with the scikit-rebate's ReliefF implementation, we decided to use the older version in which ReliefF is offered alone.}; and finally the SVM-RFE FS algorithm, which was not used in the present work due to high computational costs, as well as the SVM and GBM classifier models were implemented with the scikit-learn Python package \citep{scikit-learn}.

Moreover, the framework presents an outstanding feature: its capability of accepting new FS algorithms (either written in Python or R) added by the user, as well as new aggregation strategies and new classifier models. This is a relevant component that empowers scientific investigation by providing scalability to its methodology and allowing brand new ensemble FS setups with minimal programming overhead. All technical details can be found in the tools' GitHub repository. We note that the implemented tools currently only support FS for binary supervised classification problems, and the applicability of the algorithms for different domain data is yet to be analyzed. Nonetheless, both tools are open source, and we encourage other researchers interested in using or expanding them to contribute to their growth.

\section{Conclusions}
\label{conclusion}

In this work, considerable effort was dedicated to improving the state-of-the-art related to feature selection from high-dimensional data using ensemble approaches. We provided evidence that out Hybrid EFS in its distinct variants, but mainly the Hyb-Wx-GR-SU, is a promising solution for this purpose. Our results were not only better in terms of stability and predictive potential, but also in terms of biological plausibility. Moreover, we carried out a comprehensive experimental comparison among distinct strategies for FS, including single FS algorithms and three types of ensembles, using cancer-related transcriptome data, which may provide valuable insights for future works. We showed that the proposed Hybrid EFS approach improved upon the general performance of the well-established Heterogeneous EFS strategy, both in quantitative and in qualitative comparisons, and that the difference is even more outstanding if base FS algorithms are refined based on their performance. Thus, automatic selection of ensemble components during model training could be a promising research direction to optimize its performance for future work.

\section*{Acknowledgment}
This study was financed in part by the Coordenação de Aperfeiçoamento de Pessoal de Nível Superior - Brasil (CAPES) - Finance Code 001, Conselho Nacional de Desenvolvimento Científico e Tecnológico (CNPq), and Fundação de Amparo à Pesquisa do Estado do Rio Grande do Sul (FAPERGS).

\bibliographystyle{elsarticle-num-names}
\bibliography{references}

\end{document}